\documentclass[journal,final]{IEEEtran}
\usepackage[T1]{fontenc}
\usepackage{fancyhdr,comment,setspace,algorithm,algorithmic}
\usepackage{amsmath,amssymb,amsthm,graphicx,graphics,subfigure,epsfig,enumerate}
\newcommand{\ba}{\boldsymbol{a}}

\newcommand{\bn}{\boldsymbol{n}}
\newcommand{\bb}{\boldsymbol{b}}
\newcommand{\bSigma}{\boldsymbol{\Sigma}}
\newcommand{\bLambda}{\boldsymbol{\Lambda}}
\newcommand{\bPi}{\boldsymbol{\Pi}}
\newcommand{\bGamma}{\boldsymbol{\Gamma}}
\newcommand{\bnu}{\boldsymbol{\nu}}
\newcommand{\cF}{\mathcal{F}}

\newcommand{\cE}{\mathcal{E}}
\newcommand{\Var}{\mathrm{Var}} 
\newcommand{\Tr}{\mathrm{Tr}} 

\newcommand{\argmax}{\mathrm{argmax}} 
\newcommand{\argmin}{\mathrm{argmin}} 
\newcommand{\F}{\mathrm{F}} 

\newcommand{\bP}{{\boldsymbol P}} 
\newcommand{\bK}{{\boldsymbol K}} 
\newcommand{\bA}{{\boldsymbol A}}
\newcommand{\bB}{{\boldsymbol B}}

\newcommand{\bx}{{\boldsymbol x}}

\newcommand{\bM}{{\boldsymbol M}}
\newcommand{\bU}{{\boldsymbol U}}
\newcommand{\cT}{{\mathcal T}}

\newcommand{\bQ}{{\boldsymbol Q}}
\newcommand{\bR}{{\boldsymbol R}}

\newcommand{\bJ}{{\boldsymbol J}}

\newcommand{\bu}{{\boldsymbol u}}
\newcommand{\bv}{{\boldsymbol v}}
\newcommand{\bI}{{\boldsymbol I}}

\newcommand{\bV}{{\boldsymbol V}}
\newcommand{\bX}{{\boldsymbol X}}

\newcommand{\bW}{{\boldsymbol W}}
\graphicspath{{Graphs/}}

\title{Subspace Learning From Bits}
\author{Yuejie Chi,~\IEEEmembership{Member,~IEEE}, Haoyu Fu,~\IEEEmembership{Student Member,~IEEE} \thanks{The authors are with the Department of Electrical and Computer Engineering, The Ohio State University, Columbus, OH 43210. Email: \{chi.97, fu.436\}@osu.edu.}
\thanks{Preliminary results have been presented in part at the 2014 IEEE Global Conference on Signal and Information Processing (GlobalSIP) \cite{chi2014onebit} and the 2016 Asilomar Conference on Signals, Systems, and Computers \cite{fu2016asilomar}.} 
\thanks{This material is based upon work supported by the Air Force Summer Faculty Fellowship Program, by National Science Foundation under award number ECCS-1462191, and by the Air Force Office of Scientific Research under award number FA9550-15-1-0205. }}

\date{\today}

\theoremstyle{plain}\newtheorem{lemma}{\textbf{Lemma}}\newtheorem{theorem}{\textbf{Theorem}}
\newtheorem{prop}{\textbf{Proposition}}
\theoremstyle{definition}
 
\begin{document}

\maketitle


\begin{abstract}
Networked sensing, where the goal is to perform complex inference using a large number of inexpensive and decentralized sensors, has become an increasingly attractive research topic due to its applications in wireless sensor networks and internet-of-things. To reduce the communication, sensing and storage complexity, this paper proposes a simple sensing and estimation framework to faithfully recover the principal subspace of high-dimensional data streams using a collection of binary measurements from distributed sensors, without transmitting the whole data. The binary measurements are designed to indicate comparison outcomes of aggregated energy projections of the data samples over pairs of randomly selected directions. When the covariance matrix is a low-rank matrix, we propose a spectral estimator that recovers the principal subspace of the covariance matrix as the subspace spanned by the top eigenvectors of a properly designed surrogate matrix, which is provably accurate as soon as the number of binary measurements is sufficiently large. An adaptive rank selection strategy based on soft thresholding is also presented. Furthermore, we propose a tailored spectral estimator when the covariance matrix is additionally Toeplitz, and show reliable estimation can be obtained from a substantially smaller number of binary measurements. Our results hold even when a constant fraction of the binary measurements is randomly flipped. Finally, we develop a low-complexity online algorithm to track the principal subspace when new measurements arrive sequentially. Numerical examples are provided to validate the proposed approach.
\end{abstract}
 

\section{Introduction}
Networked sensing, where the goal is to perform complex inference using data collected from a large number of inexpensive and decentralized sensors, has become an increasingly attractive research topic in recent years due to its applications in wireless sensor networks and internet-of-things. Consider, for example, a data stream which generates a zero-mean high-dimensional data sample $\bx_t\in\mathbb{C}^{n}$ at each time $t$, and each sensor may access a portion of the data stream. Several main challenges arise when processing the high-dimensional data stream: 
\begin{itemize}
\item \textbf{Data on-the-fly:} Due to the high rate of data arrival, each data sample $\bx_t$ may not be fully stored, and computation needs to be accomplished with only one pass or a few passes \cite{babcock2002models} at the sensors to allow fast processing. 
\item \textbf{Resource constraints:} The sensors usually are power-hungry and resource-limited, therefore it is highly desirable to minimize the computational and storage cost at the sensor side, as well as the communication overheads to the fusion center by not transmitting all the data. 
\item \textbf{Dynamics:} as new data samples arrive and/or new sensors enter, it is interesting to {\em track} the changes of the {\em information flow} at the fusion center without storing all history data in a low-complexity fashion.
\end{itemize}

\subsection{Contributions of this paper}
Many practical data exhibit low-dimensional structures, such that a significant proportion of their variance can be captured in the first few principal components; and that the subspace spanned by these principal components is the recovery object of interest rather than the datasets themselves. In other words, the covariance matrix of the data $\bSigma=\mathbb{E}[\bx_t\bx_t^H]$ is (approximately) low-rank, $\mbox{rank}(\bSigma)\approx r$, where $r\ll n$. This assumption is widely applicable to data such as network traffic, wideband spectrum, surveillance, and so on. 
 
To reduce the communication, sensing and storage complexity, it is of great interest to consider one-bit sampling strategies at decentralized sensor nodes \cite{luo2005universal,ribeiro2006bandwidth,frugal2013}, where each sensor only transmits a single bit to the fusion center rather than all the data.  
Our goal in this paper is thus to design a simple yet efficient one-bit sampling strategy to faithfully retrieve the {\em principal subspace} of high-dimensional data streams with i.i.d. samples when their covariance matrices are (approximately) low-rank with provable performance guarantees. We focus on the scenario where each distributed sensor may only access a subset of the whole data, process them locally, and transmit only a single bit to the fusion center, who will estimate the principal subspace without referring to the original data. It is of interest to reduce the amount of measurements required to communicate to the fusion center while keeping the number of sensors as small as possible to allow faithful recovery of the principal subspace.

In the proposed one-bit sampling scheme, each sensor is equipped with a pair of  vectors composed of i.i.d. complex standard Gaussian entries, called {\em sketching vectors}. At the sensing stage, it compares the energy projections of the data seen at the sensor onto the pair of sketching vectors respectively, and transmits a single bit indicating which of the two energy projections is larger. This is equivalent to comparing the energy projection of a {\em sample covariance matrix} onto two randomly selected rank-one subspaces. A key observation is that as long as the number of samples seen at the sensor is not too small (which we characterize theoretically), the comparison outcome will be exactly the same as if it is performed on the {\em covariance matrix} or its best low-rank approximation. By only transmitting the comparison outcome rather than the actual energy measurements, the communication overhead is minimized to a single bit which is more robust to communication channel errors and outliers. Moreover, as will be shown, the energy projections can be computed extremely simple without storing the history data samples, and are always nonnegative, making them suitable for wideband and optical applications. 

At the fusion center, the sketching vectors are assumed known, which is a standard assumption for decentralized estimation \cite{baron2009distributed}. When the covariance matrix is exactly low-rank, we propose a spectral estimator, which estimates the principal subspace as the subspace spanned by the top eigenvectors of a carefully designed surrogate matrix using the collected binary measurements, which can be easily computed via a truncated eigenvalue decomposition (EVD) with a low computational complexity. We show that, assuming all the bit measurements are exactly measuring the covariance matrix, the estimate of the principal subspace is provably accurate as long as the number of bits is on the order of $nr^3\log n$, when the sketching vectors are composed of i.i.d. standard complex Gaussian entries. When the rank is not known a priori, we devise a soft-thresholding strategy to adaptively select the rank, and show it obtains a similar performance guarantee. Furthermore, we developed a memory-efficient algorithm to online update the principal subspace estimate when the binary measurements arrive sequentially, which can be implemented with a memory requirement on the order of the size of the principal subspace rather than that of the covariance matrix.


In many applications concerning (power) spectrum estimation, such as array signal processing and cognitive radios, the covariance matrix of the data can be  modeled as a low-rank {\em Toeplitz} matrix \cite{li2016offgrid}. It is therefore possible to further reduce the required number of bit measurements by exploiting the Toeplitz constraint. We propose to apply the spectral estimator to the projection of the above designed surrogate matrix to its nearest Toeplitz matrix in the Frobenius norm. When the covariance matrix is rank-one, it provably admits accurate estimation of the principal subspace as soon as the number of bit measurements is on the order of $\log^4 n$. In contrast to the scenario when the Toeplitz constraint is not explored, this eliminates the linear dependency with $n$, thus the sample complexity is greatly reduced. Numerical simulations also suggest the algorithm works well even in the low-rank setting. Finally, our results continue to hold even when a constant fraction of the binary measurements is randomly flipped.

\subsection{Related work}

Estimating the principal subspace of a high-dimensional data stream from its sparse observations has been studied in recent years, but most existing work has been focused on recovery of the data stream \cite{chi2012petrels,Balzano-2010}. Recently, sketching has been promoted as a dimensionality reduction method to directly recover the {\em statistics} of the data \cite{gilbert2012sketched}\nocite{dasarathy2015sketching,Leus2011,chi2016kronecker,chen2013exact,chen2014icassp}--\cite{chen2014isit}. The proposed framework in this paper is motivated by the covariance sketching scheme in \cite{chen2013exact,chen2014icassp,chen2014isit}, where a quadratic sampling scheme is designed for low-rank covariance estimation. It is shown in \cite{chen2013exact} that a number of real-valued quadratic (energy) measurements on the order of $nr$ suffices to exactly recover rank-$r$ covariance matrices via nuclear norm minimization, assuming the measurement vectors are composed of i.i.d. sub-Gaussian entries. However, transmitting these energy measurements with high precision may cause unwanted overhead and require estimating the noise level in practice. 



Distributed estimation of a {\em scalar-valued} parameter from the one-bit quantization of its noisy observations has been considered in  \cite{luo2005universal,ribeiro2006bandwidth,luo2005isotropic,xiao2006distributed}. Recently, one-bit compressed sensing \cite{boufounos20081}\nocite{jacques2013robust,plan2013one,plan2013robust,haupt2011robust}--\cite{gupta2010sample} and one-bit matrix completion \cite{davenport20141,cai2013max} have generalized this to the estimation of vector-valued parameters such as {\em sparse vectors} and {\em low-rank matrices}, where they aim to recover the signal of interest from the signs of its random linear measurements. Our work is related to one-bit matrix completion as we also consider low-rank structures of the covariance matrices, but differs in several important aspects. First, unlike existing work, our binary measurements are not constructed directly from the low-rank covariance matrix, but rather a sample covariance matrix, therefore we need to carefully justify when the binary measurements accurately reflect the characteristic of the true covariance matrix. Second, the measurement operators with respect to the covariance matrix take the form of the difference of two rank-one matrices, designed to allow a low-complexity implementation, which leads to very different results from existing ones that assume i.i.d. entries \cite{plan2013robust}. Third, we propose simple spectral estimators for reconstructing the principal subspace of low-rank covariance matrices, and demonstrate both analytically and through simulations that it obtains similar performance as the more expensive convex programs using trace norm minimization \cite{plan2013robust}. Finally, the spectral estimator can be further tailored to the case of low-rank Toeplitz covariance matrices.



Distributed wideband spectrum sensing for cognitive radios is an appealing and motivating application \cite{Romero2013}. It is recently proposed to estimate the power spectral density of wideband signals via least-squares estimation from sub-Nyquist samples \cite{ariananda2012compressive}. The frugal sensing framework \cite{frugal2013} considered the same estimation problem using one-bit measurements based on comparing the average signal power within a band of interest against a pre-determined or adaptively-set threshold \cite{Mehanna2013}. Their algorithm is based on linear programming and may explore parametric representations of the power spectral density. Our work is different from \cite{frugal2013,Mehanna2013} in several aspects. Instead of comparing the average signal power against a threshold which introduces the additional issue of how to set the threshold, we compare the average signal power between two different bands of interest and therefore do not need to set any threshold. Our algorithm explores the low-rank property of the power spectral density rather than its parametric representation, and does not require knowing the noise statistics but explores the concentration phenomenon of random matrices.

Finally, the paper \cite{mroueh2013quantization} studied one-bit phase retrieval, an extension of the one-bit compressed sensing with phaseless measurements. Despite different motivations and applications, our algorithm subsumes the scenario in \cite{mroueh2013quantization} as a special case when the covariance matrix is assumed rank-one.




\subsection{Organization of this paper and notations}
The rest of this paper is organized as follows. Section~\ref{formulation} describes the proposed 1-bit sampling framework and formulates the principal subspace estimation problem. 
Section~\ref{performance} presents the proposed spectral estimators and their performance guarantees. Section~\ref{online} presents an online algorithm to track the low-dimensional principal subspace with sequential bit measurements. Numerical examples are given in Section~\ref{numerical}. Finally, we conclude in Section~\ref{conclusion} and outline some future directions. Additional proofs are provided in the appendix.
 
Throughout this paper, we use boldface letters to denote vectors and matrices, e.g. $\ba$ and $\bA$. The Hermitian transpose of $\ba$ is denoted by $\ba^H$, and $\| \bA\|$, $\|\bA\|_{\mathrm{F}}$, $\|\bA\|_*$, $\Tr(\bA)$ denote the spectral norm, the Frobenius norm, the nuclear norm, and the trace of the matrix $\bA$, respectively. Denote 
$$\mathcal{T} (\bA ) =\argmin_{\boldsymbol{T}}\left\|\boldsymbol{T} - \bA\right\|_{\F}\quad \mbox{s.t.}\quad \boldsymbol{T}~\text{is Toeplitz}, $$ 
as the linear projection of $\bA$ to the subspace of Toeplitz matrices, and $\mathcal{T}^{\perp} (\bA  ) = \bA -\mathcal{T}(\bA )  $. Define the inner product between two matrices $\bA,\bB$ as $\langle \bA,\bB\rangle = \Tr(\bB^H\bA)$. If $\bA$ is positive semidefinite (PSD), then $\bA \succeq 0$. The expectation of a random variable $a$ is written as $\mathbb{E}[a]$. 

\section{One-Bit Sampling Strategy}\label{formulation}
Let $\{\bx_t \}_{t=1}^{\infty}\in\mathbb{C}^n$ be a data stream with zero-mean $\mathbb{E}[\bx_t]= 0$ and the covariance matrix $\bSigma=\mathbb{E}[\bx_t\bx_t^H]$. In this section we describe the distributed one-bit sampling framework for estimating the principal subspace of $\bSigma$ based on comparison outcomes of aggregated energy projections from each sensor, as summarized in Algorithm~\ref{one_bit_PCA}.
\begin{algorithm}[hp]
\caption{One-Bit Sampling Strategy}\label{one_bit_PCA}
\textbf{Input:} A data stream $\{ \bx_t\}_{t=1}^{\infty}$;
\begin{algorithmic}[1]
\FOR{each sensor $i=1,\ldots, m$}
\STATE Randomly choose two sketch vectors $\ba_i\in\mathbb{C}^n$ and $\bb_i\in\mathbb{C}^n$ with i.i.d. Gaussian entries;
\STATE Sketch an arbitrary substream indexed by $\{\ell_t^i\}_{t=1}^T$ with two energy measurements $ |  \langle \boldsymbol{a}_{i}, \boldsymbol{x}_{\ell_{t}^i}  \rangle  |^2$ and $ |  \langle \boldsymbol{b}_{i}, \boldsymbol{x}_{\ell_{t}^i}  \rangle  |^2$, and transmit a binary bit to the fusion center:  
\begin{equation*}
 y_{i,T} = \mbox{sign}\left(\frac{1}{T} \sum_{t=1}^T  |  \langle \ba_{i}, \bx_{\ell_t^i}  \rangle  |^2- \frac{1}{T} \sum_{t=1}^T |  \langle \bb_{i}, \bx_{\ell_t^i}  \rangle  |^2 \right) . 
  \end{equation*}
\ENDFOR
\end{algorithmic}
\end{algorithm}

Consider a collection of $m$ sensors that are deployed distributively to measure the data stream. Each sensor can access either a portion or the complete data stream. At the $i$th sensor, define a pair of sketching vectors $\ba_i\in\mathbb{C}^n$ and $\bb_i\in\mathbb{C}^n$, $1\leq i\leq m$, where their entries are i.i.d. standard complex Gaussian $\mathcal{CN}(0,1)$. Without loss of generality, we assume the $i$th sensor has access to the samples in the data stream indexed by $\mathcal{L}_i = \{\ell_t^i\}_{t=1}^T$ of the same size $T$. Each $i$th sensor processes the samples locally, namely, for the data sample $\bx_{\ell_t^i}$, it takes two quadratic (energy) measurements given below:
 \begin{equation}
 u_{i,t} = |\langle \ba_i, \bx_{\ell_t^i} \rangle |^2, \quad v_{i,t} = |\langle \bb_i,  \bx_{\ell_t^i} \rangle |^2, 
 \end{equation}
which are nonnegative and can be measured efficiently in high frequency applications at a linear complexity using energy detectors. These quadratic measurements are then averaged over the $T$ samples to obtain
\begin{align*}
U_{i,T} &= \frac{1}{T}\sum_{t=1}^T u_{i,t}=\ba_i^H \bSigma_{i,T} \ba_i, \\
V_{i,T} & = \frac{1}{T}\sum_{t=1}^T v_{i,t} =\bb_i^H \bSigma_{i,T} \bb_i,
\end{align*}
where 
\begin{equation}
\bSigma_{i,T} = \frac{1}{T} \sum_{t=1}^T \bx_{\ell_t^i} \bx_{\ell_t^i}^H
\end{equation} 
is the sample covariance matrix seen by the $i$th sensor. It is clear that $U_{i,T}= \frac{T-1}{T}U_{i,T-1} + \frac{1}{T} u_{i,T} $, and similarly $V_{i,T}$, can be updated recursively without storing all the history data. At the end of the $T$ samples, the $i$th sensor compares the average energy projections $U_{i,T}$ and $V_{i,T}$, and transmit to the fusion center a single bit $y_{i,T}$ indicating the outcome:
\begin{equation}\label{sample_sign}
y_{i,T} = \left\{\begin{array}{ll}
1, & \mbox{if} ~ U_{i,T}  > V_{i,T} \\
-1, & \mbox{otherwise}
\end{array} \right. .
\end{equation}
The communication overhead is minimal since only the binary measurement is transmitted rather than the original data stream. It is also straightforward to see that each sensor only needs to store two scalars, $U_{i,T}$ and $V_{i,T}$. More concretely, define $\bW_i=\ba_{i}\ba_{i}^H -  \bb_{i} \bb_{i}^H$, we can write \eqref{sample_sign} as
\begin{equation}\label{sample_sign_Wi}
y_{i,T} = \mbox{sign}\left(\langle \bW_i, \bSigma_{i,T} \rangle \right) ,
\end{equation}
where $\mbox{sign}(\cdot)$ is the sign function. Intuitively, \eqref{sample_sign_Wi} can be interpreted as comparing the energy projection of $\bSigma_{i,T}$ onto two randomly selected rank-one subspaces. Finally, to model potential errors occurred during transmission, we assume each bit has an independent {\em flipping probability} of $0\leq p<1/2$, and the received bit at the fusion center from the $i$th sensor is given as
\begin{equation}\label{noisy_1bit}
z_{i,T} = y_{i,T} \cdot \epsilon_i, \quad 1\leq i\leq m,
\end{equation}
where $\mathbb{P}(\epsilon_i = -1)=p$ and $\mathbb{P}(\epsilon_i = 1)=1- p$ are i.i.d. across sensors.

As we're interested in the covariance matrix $\bSigma$, and note that the sample covariance matrix $\bSigma_{i,T}$ converges to $\bSigma$ as $T$ approaches infinity, the binary measurement at the $i$th sensor also approaches quickly to the following,
\begin{equation}\label{sign_Wi}
 y_i = \mbox{sign}\left(\langle \bW_i, \bSigma \rangle \right),
\end{equation}
as if it is measuring the true covariance matrix $\bSigma$. In fact, as we'll show in Theorem~\ref{all_bits_sample}, $y_i$ and $y_{i,T}$ start to agree very fast for $T$ much smaller than $n$. 

For simplicity we assume $\bSigma$ is an exactly rank-$r$ PSD matrix, where $r\ll n$. The extension to approximately low-rank case will be discussed shortly. Let 
\begin{align}
\bSigma &=  \sum_{k=1}^r \lambda_k \bu_k\bu_k^H= \bU\bLambda\bU^H ,
\end{align}
where $\bU = [\bu_1,\bu_2,\ldots,\bu_r]$ are the top-$r$ eigenvectors of $\bSigma$ and $\{\lambda_k\}_{k=1}^r$ are the top-$r$ eigenvalues with $\lambda_1 \geq \lambda_2 \geq \cdots \geq \lambda_r$. We further define $\bv_k$, $k=1,\ldots, n-r$ as the basis vectors spanning the complement of $\bU$. Apparently not all information about $\bSigma$ can be recovered, for example, the sign measurements are {\em invariant} to scaling of the data samples, and therefore, scaling of the covariance matrix. Our goal in this paper is to recover the principal subspace spanned by $\bU\in\mathbb{C}^{n\times r}$ from the collected binary measurements.

\subsection{How large does $T$ need to be?}



In the following proposition, whose proof can be found in Appendix~\ref{proof_single_bit_sample}, we establish the sample complexity of $T$ to guarantee $y_i = y_{i,T}$ if the data stream follows a Gaussian model $\bx_t\sim\mathcal{CN}(\mathbf{0},\bSigma)$ with i.i.d. samples. 

\begin{prop}\label{single_bit_sample} Let $0<\delta\leq 1$. Assume $\bx_t$ are i.i.d. Gaussian satisfying $\bx_t\sim\mathcal{CN}(\mathbf{0},\bSigma)$. Then
$ \mathbb{P} \left[ y_{i,T} \neq  y_i \right] \leq \delta $
as soon as $T > c \frac{\mathrm{Tr}(\bSigma)}{\|\bSigma\|_{\mathrm{F}}}\log^2(1/\delta) $ for some sufficiently large constant $c$.
\end{prop}




In order to guarantee that all $m$ bits are accurate, we need to further apply a union bound to Proposition~\ref{single_bit_sample}, which yields the following theorem.
\begin{theorem}\label{all_bits_sample} Let $\bx_t$ be i.i.d. $\bx_t\sim\mathcal{CN}(0, \bSigma)$. Let $0<\delta\leq 1$. With probability at least $1-\delta$, all binary measurements are exact, i.e. $y_{i,T}=y_i$ for $i=1,\ldots, m$ given that the number of samples observed by each sensor satisfies
$$T > c \frac{\mathrm{Tr}(\bSigma)}{\|\bSigma\|_{\mathrm{F}}}\log^2\left(\frac{m}{\delta}\right) $$ 
for some sufficiently large constant $c$.
\end{theorem}

It is worth emphasizing that Theorem~\ref{all_bits_sample} holds for any fixed covariance matrix $\bSigma$. The term $\Tr(\bSigma)/\|\bSigma\|_{\F}$ measures the ``effective'' rank of $\bSigma$, as for PSD matrices with fast spectral decays this term will be small \cite{Vershynin2012}. If $\mbox{rank}(\bSigma)=r$, then $\Tr(\bSigma) \leq \sqrt{r}\| \bSigma \|_{\mathrm{F}}$. As soon as $T$ is on the order of $\sqrt{r}\log^2 m$ all bit measurements are accurate with high probability, which only depends on the number of sensors (which in turn depend on the ambient dimension $n$ as will be seen from Theorem~\ref{theorem_main}) logarithmically. 

\subsection{Extension to approximate low-rank covariance matrices} 
When $\bSigma$ is only approximately low-rank, denote its best rank-$r$ approximation as $\bSigma_r = \argmin_{\mbox{rank}(\bA)=r}\|\bSigma-\bA\|_{\F}$, then if 
\begin{equation} \label{ext_approximate_lowrank}
\mbox{sign}\left(\langle \bW_i, \bSigma \rangle \right) = \mbox{sign}\left(\langle \bW_i, \bSigma_r \rangle \right), \forall i=1,\ldots, m, 
\end{equation}
holds, combined with Theorem~\ref{all_bits_sample}, our framework can be applied to recover the $r$-dimensional principal subspace of $\bSigma$, by treating the bits as measurements of $\bSigma_r$. Fortunately, \eqref{ext_approximate_lowrank} holds with high probability as long as $\|\bSigma-\bSigma_r\|_*/\|\bSigma\|_{\F}\leq c \log(m)/n$ is sufficiently small. The interested readers are referred to Appendix~\ref{approx_lr} for the details.

\section{Principal Subspace Estimators and Performance Guarantees}\label{performance}

In this section, we first develop a spectral estimator when $\bSigma$ is a low-rank PSD matrix based on truncated EVD when the rank is known exactly; and then we develop a soft-thresholding strategy to adaptively select the rank when it's unknown. Finally, we tailor the spectral estimator to the case when $\bSigma$ is a low-rank Toeplitz PSD matrix. For the rest of this section, we assume the binary measurements $y_{i,T}=y_i$ and $z_i = y_i \cdot \epsilon_i$ in \eqref{noisy_1bit} for all $i=1,\cdots,m$.

\subsection{The spectral estimator}\label{algorithm}
We propose an extremely simple and low-complexity spectral estimator whose complexity amounts to computing a few top eigenvectors of a carefully designed surrogate matrix. To motivate the algorithm, consider the special case when $\bSigma=\bnu \bnu^H$ is a rank-one matrix with $\|\bnu\|_{2}=1$. A natural way to recover $\bnu$ is via the following:
\begin{align}\label{optimization_view}
& \max_{\bnu:\|\bnu\|_2=1 } \; \frac{1}{m}\sum_{i=1}^m z_i \left\langle \bW_i,  \bnu\bnu^H \right\rangle,
\end{align}
which aims to find a rank-one matrix $\bnu\bnu^H$ that agrees with the measured signs as much as possible. Since \eqref{optimization_view} is equivalent to
\begin{align*}
&  \max_{\bnu:\|\bnu\|_2=1 }   \bnu^H \left( \frac{1}{m}\sum_{i=1}^m z_i\bW_i \right) \bnu , 
\end{align*}
its solution is the top eigenvector of the surrogate matrix:
\begin{equation}\label{surrogate_sample}
\boldsymbol{J}_{m}= \frac{1}{m}\sum_{i=1}^m z_i \bW_i = \frac{1}{m} \sum_{i=1}^m \epsilon_i \mbox{sign}(\langle \bW_i, \bSigma \rangle) \bW_i.
\end{equation}

More generally, when $\bSigma$ is rank-$r$, we recover its principal subspace as the subspace spanned by the top-$r$ eigenvectors $\widehat{\bU}\in\mathbb{C}^{n\times r}$ of the surrogate matrix $\bJ_m$ in \eqref{surrogate_sample}. This procedure is denoted as the spectral estimator. 



\subsection{Sample complexity of the spectral estimator}

\begin{figure}[t]
\begin{center}
\includegraphics[width=0.4\textwidth]{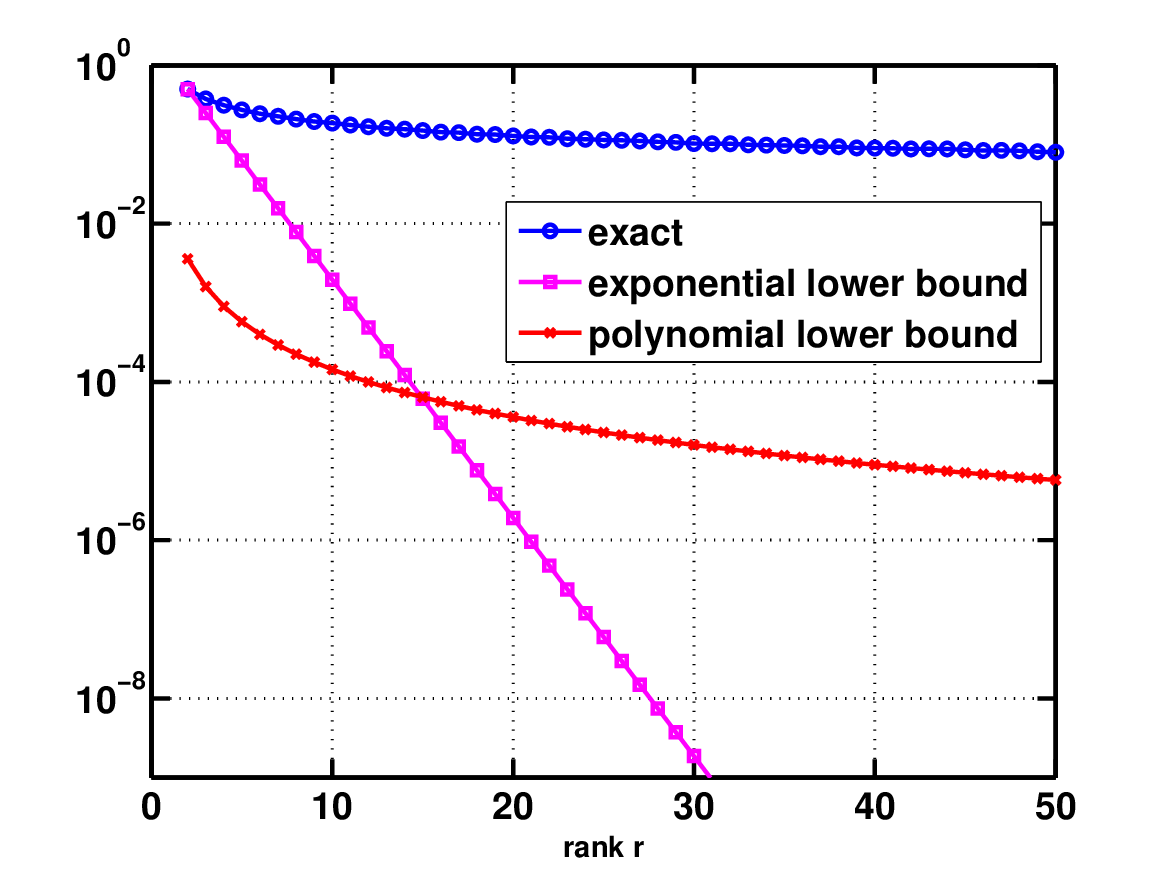}
\caption{The derived lower bounds in Lemma~\ref{evd_expectedJ} compared with its exact values when all $\lambda_k=1$ for different ranks.}\label{lowerbound}
\end{center}
\end{figure}

We establish the performance guarantee of the proposed spectral estimator by showing that the principal subspace of $\bSigma$ can be accurately estimated using $\bJ_m$ as soon as $m$ is sufficiently large. This is accomplished in two steps. Define $\bJ = \mathbb{E}[\bJ_m]$. We first show that the principal subspace of $\bJ$ is the same as that of $\bSigma$, and then show that $\bJ_m$ is concentrated around $\bJ$ for sufficiently large $m$. The following lemma accomplishes the first step.
\begin{lemma}\label{evd_expectedJ}The principal subspace of ${\bJ}$ is the same as that of $\bSigma$ with $\mbox{rank}( {\bJ})=r$. For $k=1,\ldots, r$,
\begin{align}
1\geq \frac{\bu_k^H  \bJ \bu_k}{(1-2p)} &\geq  \max\left\{  \left(\frac{1}{1+\kappa(\bSigma)}\right)^{r-1}, \frac{1}{9r}e^{-\kappa(\bSigma)} \right\}, \label{principal}
\end{align}
and for $k=1,\ldots, n-r$,
\begin{align}
\bv_k^H  {\bJ} \bv_k &= 0 , \label{minor}
\end{align}
where $\kappa(\bSigma) = \lambda_1/\lambda_r$ is the conditioning number of $\bSigma$. When $r=1$, the right-hand side of \eqref{principal} equals one.
\end{lemma}
The proof is provided in Appendix~\ref{proof_evd_expectedJ}. Lemma~\ref{evd_expectedJ} establishes that $\bJ$ and $\bSigma$ share the same principal subspace, but their eigenvectors may still differ. Following Lemma~\ref{evd_expectedJ}, the spectral gap between the $r$th eigenvalue and the $(r+1)$th eigenvalue (which is zero) of ${\bJ}$ is at least 
$$\alpha: = (1-2p) \max\left\{  \left(\frac{1}{1+\kappa(\bSigma)}\right)^{r-1}, \frac{1}{9r}e^{-\kappa(\bSigma)}   \right\} , $$
 where the first term (exponential in $r$) is tighter when $r$ is small while the second term (polynomial in $r$) is tighter when $r$ is large. Indeed, when $k=1,\ldots,r $, $\bu_k^H \bJ \bu_k$ only depends on $\{\lambda_k\}_{k=1}^r$ and can be computed exactly once they are fixed. Fig.~\ref{lowerbound} plots the derived lower bounds and the exact values of $\bu_k^H \bJ \bu_k$ assuming all $\lambda_k=1$. It can be seen that the polynomial bound on the order of $1/r$ is rather accurate except the leading constant when $r$ is large. Moreover, from Fig.~\ref{lowerbound} it confirms that although $\bJ$ preserves the principal subspace of $\bSigma$, it does not preserve the eigenvectors and eigenvalues.


Next, we show that for sufficiently large $m$, the matrix $\bJ_m$ is close to its expectation ${\bJ}$. We have the following lemma whose proof can be found in Appendix~\ref{proof_J_concentration}.
\begin{lemma}\label{J_concentration}
 Let $0<\delta<1$. Then with probability at least $1-\delta$, we have that 
$$ \| \bJ_m -  {\bJ} \| \leq  \sqrt{\frac{c_1n}{m}\log\left(\frac{2n}{\delta} \right)}, $$
where $c_1$ is an absolute constant.
\end{lemma}

 
Our main theorem then immediately follows by applying an improvement of the Davis-Kahan sin-Theta theorem \cite{lei2013consistency} in Lemma~\ref{davis_kahan}, as given below.
\begin{theorem} \label{theorem_main}Let $0<\delta<1$. With probability at least $1-\delta$, there exists an $r\times r$ orthonormal matrix $\bQ$ such that  
\begin{align*}
& \left\| \widehat{\bU} - \bU\bQ \right \|_{\mathrm{F}}  \leq \alpha^{-1} \sqrt{\frac{c_1nr}{m}\log\left(\frac{2n}{\delta} \right)} \\
& \leq \frac{\min \left\{  \left( 1+\kappa(\bSigma)\right)^{r-1}, 9e^{\kappa(\bSigma)} r\right\}}{(1-2p)} \sqrt{\frac{c_1nr}{m}\log\left(\frac{2n}{\delta} \right)},
\end{align*}
where $c_1$ is an absolute constant.
\end{theorem}

When $\kappa(\bSigma)$ is small and $r$ is moderate, $\alpha$ scales as $1/r$ and we have that as soon as the number of binary measurements $m$ exceeds the order of $nr^3 \log n$, it is sufficient to recover to recover the principal subspace spanned by the columns of $\bU$ with high accuracy. Since there are at least $nr$ degrees of freedom to describe $\bU$, our bound is near-optimal up to a polynomial factor with respect to $r$ and a logarithmic factor with respect to $n$. It is worth emphasizing that our result indicates that the order of required binary measurements is much smaller than the ambient dimension of the covariance matrix, and even comparable to the sample complexity for low-rank matrix recovery using real-valued measurements \cite{RecFazPar07}. Furthermore, the reconstruction is robust to random flipping errors, as long as $p<1/2$. In fact, the error scales inverse proportionally to the expectation of correct transmission.

\subsection{Adaptive Rank Selection via Soft-thresholding}\label{convex}

The performance guarantee of the spectral estimator in Theorem~\ref{theorem_main} requires perfect knowledge of the rank, which may not be readily available. One possible strategy is to soft-threshold the eigenvalues of $\bJ_m$ to select a proper dimension of the principal subspace. We motivate this choice as the solution to a regularized convex optimization problem whose analysis sheds light on how to select the threshold. To begin with, following the rationale of \eqref{optimization_view}, we may seek to find a  low-rank PSD matrix $\widehat{\bSigma}$ that obeys
\begin{align*}
\widehat{\bSigma}&=\argmax_{\bSigma\succeq 0} \frac{1}{m}\sum_{i=1}^m z_i \langle \bSigma, \bW_i \rangle = \argmax_{\bSigma\succeq 0}\langle \bSigma, \bJ_m\rangle, \\
&  \quad\quad\mbox{s.t.}\quad  \mbox{rank}(\bSigma)\leq r,\quad\; \| \bSigma\|_{\mathrm{F}}\leq 1.
\end{align*}
However, this formulation is non-convex due to the rank constraint. We therefore consider the following convex relaxation by relaxing the rank constraint by trace minimization and regularizing the norm of $\bSigma$, yielding:
\begin{align}\label{cvx_regularized}
\widehat{\bSigma} &=\argmin_{\bSigma\succeq 0}\; \left\| \bSigma-\bJ_m \right\|_{\F}^2+\lambda\Tr\left(\bSigma\right),
\end{align}
where $\lambda>0$ is the regularization parameter. The above problem \eqref{cvx_regularized} admits a closed-form solution \cite{cai2010singular}. Let the EVD of $\bJ_m$ be given as $\bJ_m = \sum_{k=1}^n \widehat{\lambda}_k \widehat{\bu}_k\widehat{\bu}_k^H$, where the $\widehat{\lambda}_k$'s are ordered in the descending order. Then $\widehat{\bSigma}= \sum_{k=1}^{n}\widehat{\eta}_k \widehat{\bu}_k\widehat{\bu}_k^H$, where $\widehat{\eta}_k =\max\{ 0, \widehat{\lambda}_k-\lambda\}$. Therefore, like the spectral estimator, the estimated principal space are spanned by the eigenvectors of $\widehat{\bSigma}$ corresponding to the nonzero eigenvalues, where the rank is now set adaptively via soft-thresholding by $\lambda$. The performance of \eqref{cvx_regularized} can be bounded by the following lemma from \cite[pp. 1267-1268]{li2016offgrid}.
\begin{lemma}\label{lemma:opt_condition}
Set $\lambda \geq \| \bJ - \bJ_m \|$, then the solution $\widehat{\bSigma}$ to \eqref{cvx_regularized} satisfies:
$$\|\widehat{\bSigma}- {\bJ} \|_{\F} \leq  12\lambda\sqrt{2r}. $$
\end{lemma}
Combined with Lemma~\ref{J_concentration}, if we set $\lambda = \sqrt{\frac{c_1n}{m}\log\left(\frac{2n}{\delta} \right)}$, we then have
\begin{equation} \label{cvx_guarantee}
\left\|\widehat{\bSigma}- {\bJ} \right\|_{\F} \leq  c\sqrt{\frac{nr}{m}\log\left(\frac{2n}{\delta} \right)},
\end{equation}
for some constant $c$. Therefore, by the Davis-Kahan theorem, we again obtain a performance guarantee on the recovered principal subspace that is qualitatively similar to that in Theorem~\ref{theorem_main}, except that now the spectral estimator employs an adaptive strategy to select the rank via soft-thresholding. 

\textbf{Remark:} It is worthwhile to pause and comment on the difference between the convex regularized algorithm \eqref{cvx_regularized} with the algorithm proposed by Plan and Vershynin in \cite{plan2013robust} for one-bit matrix completion, which can be written as 
\begin{align}\label{algorithm} 
\max_{\bSigma \succeq 0} \sum_{i=1}^m y_i \langle \bW_i, \bSigma \rangle,  \; \mbox{s.t.} \; \;\|\bSigma\|_{\mathrm{F}} \leq 1, \; \Tr(\bSigma) \leq \sqrt{r}.
\end{align}
Indeed, the two algorithms are essentially the same and \eqref{cvx_guarantee} is also applicable to \eqref{algorithm}\footnote{The analysis is straightforward by slightly adapting the arguments, therefore we omit the details.}. However, the analysis in \cite{plan2013robust} assumes that $\bW_i$'s are composed of i.i.d. Gaussian entries, where $\bJ$ can be shown as a scaled version of $\bSigma$, so that the performance bound in \eqref{cvx_guarantee} guarantees that one can recover the {\em low-rank covariance matrix} up to a scaling difference as soon as $m$ is on the order of $nr\log n$. Unfortunately, as in our sampling scheme, due to the dependence of the entries of $\bW_i$, $\bJ$ is no longer a scaled variant of $\bSigma$ (as verified in Fig.~\ref{lowerbound}), it is not clear whether it is possible to recover the covariance matrix in a straightforward manner. Nonetheless, we evaluate the performance of \eqref{algorithm} numerically in the Section~\ref{numerical} for principal subspace estimation, and show it is comparable to that of the spectral estimator, while incurring a much higher computational cost.

\subsection{Rank-One Toeplitz Subspace Estimation}\label{section:main}
In many applications, the covariance matrix $\bSigma$ can be modeled as a low-rank Toeplitz PSD matrix, and it is desirable to further reduce the sampling complexity by exploiting the Toeplitz constraint. Denote $\cT(\bJ_m)$ as the projection of $\bJ_m$ onto Toeplitz matrices, we can show that it concentrates around the Toeplitz matrix $\cT(\bJ)$ at a rate much faster than Lemma~\ref{J_concentration}, as soon as $m$ scales poly-logarithmically with respect to $n$.
\begin{lemma}\label{lemma:concentration}
With probability at least $1-n^{-9}$, we have
$$\left\| \mathcal{T}(\bJ) - \mathcal{T}\left(\bJ_m\right) \right\| \leq c_2\cdot\frac{\log^2n}{\sqrt{m}}, $$
 where $c_2$ is some constant.
\end{lemma}
The proof can be found in Appendix~\ref{proof_lemma_concentration}. When $\bSigma$ is rank-one, by Lemma~\ref{evd_expectedJ}, $\bJ = (1-2p) \bu_1\bu_1^H$ where $\bu_1$ is the top eigenvector of $\bSigma$, therefore $\bJ$ is also rank-one and Toeplitz. Denote the top eigenvector of $\cT(\bJ_m)$ as $\widehat{\bu}_1$, combining Lemma~\ref{lemma:concentration} and Lemma~\ref{davis_kahan}, we have the following theorem.
\begin{theorem}\label{theorem_toeplitz}
Assume $\mbox{rank}(\bSigma)=1$. With probability at least $1-n^{-9}$, there exists $\theta\in[0,2\pi)$ such that
$$\left\|\widehat{\bu}_1 - e^{j\theta} \bu_1 \right\| \leq    \frac{ c_3}{(1-2p)} \sqrt{\frac{\log^4 n}{m}}$$
for some constant $c_3$.
\end{theorem} 
Therefore, Theorem~\ref{theorem_toeplitz} indicates that for the rank-one case, $\bu_1$ can be accurately estimated as long as $m$ scales on the order of $\log^4 n$, and the reconstruction is robust to random flipping errors in the received bits. This is a much smaller sample complexity compared with Theorem~\ref{theorem_main} when the Toeplitz structure is not exploited, which requires $m$ scales at least linearly with respect to $n$. 

\textbf{Remark:} In the general low-rank case, $\bJ$ may not be Toeplitz even when $\bSigma$ is Toeplitz, which prohibits us from obtaining the performance guarantee. However, the simulation suggests that the spectral estimator continues to perform well in the low-rank case, whose investigator we leave to future work.

%

\section{Subspace Tracking with Online Binary Measurements} \label{online}
In this section, we develop an online subspace estimation and tracking algorithm for the fusion center to update the principal subspace estimate of the low-rank covariance matrix when new binary measurements arrive sequentially. This is particularly useful when the fusion center is {\em memory limited} since the proposed algorithm only requires a memory space on the order of $nr$, which is the size of the principal subspace.

 Essentially we need to update the principal subspace of $\bJ_m$ from that of $\bJ_{m-1}$ given a rank-two update as follows:
\begin{equation}\label{reweight}
\bJ_m =\frac{m-1}{m}\bJ_{m-1} + \frac{y_m}{m} \left(\ba_m\ba_m^H - \bb_m\bb_m^H\right)
\end{equation}
We can rewrite \eqref{reweight} using more general notations as
\begin{equation}\label{rescaled}
\bJ_m = \eta_m \bJ_{m-1} + \bK_m\bLambda_m \bK_m^H,
\end{equation}
where \eqref{reweight} can be obtained from \eqref{rescaled} by letting $\eta_m =\frac{m-1}{m}$, $\bK_m = [\ba_m , \bb_m ]\in\mathbb{C}^{n\times 2}$, and $\bLambda_m = \mbox{diag}([y_m/m, -y_m/m])$. Note that it might be of interest to incorporate an additional discounting factor on $\eta_m$ to emphasize the current measurement, by letting $\eta_m$ to take a smaller value, as done in \cite{brand2002incremental}.

Assume the EVD of $\bJ_{m-1}$ can be written as $\bJ_{m-1}=\bU_{m-1}\bPi_{m-1} \bU_{m-1}^H$ where $\bU_{m-1}\in\mathbb{C}^{n\times r}$ is orthonormal and $\bPi_{m-1}\in\mathbb{R}^{r\times r}$ is diagonal. The goal is to find the best rank-$r$ approximation of $\bJ_m$ by updating $\bU_{m-1}$ and $\bPi_{m-1}$.

We develop a fast rank-two update of the EVD of a symmetric matrix by introducing necessary modifications of the incremental SVD approach in \cite{brand2002incremental,brand2006fast}. A key difference from \cite{brand2002incremental,brand2006fast} is that we do not allow the size of the principal subspace to grow, which is fixed as $r$. In the update we first compute an expanded principal subspace of rank $(r+2)$ and then only keep its $r$ largest principal components. 

\begin{figure*}[tp]
\begin{center}
\begin{tabular}{cc}
\includegraphics[width=0.4\textwidth]{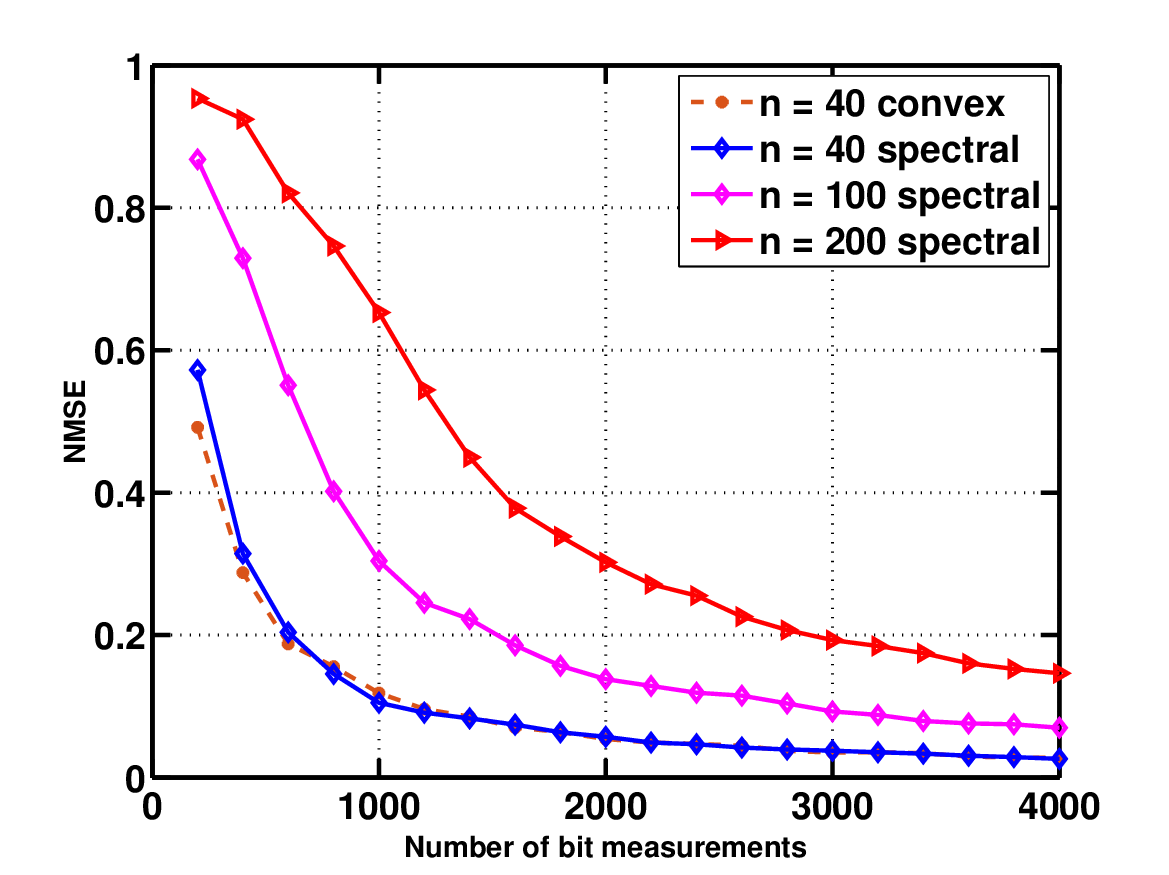} &\includegraphics[width=0.4\textwidth]{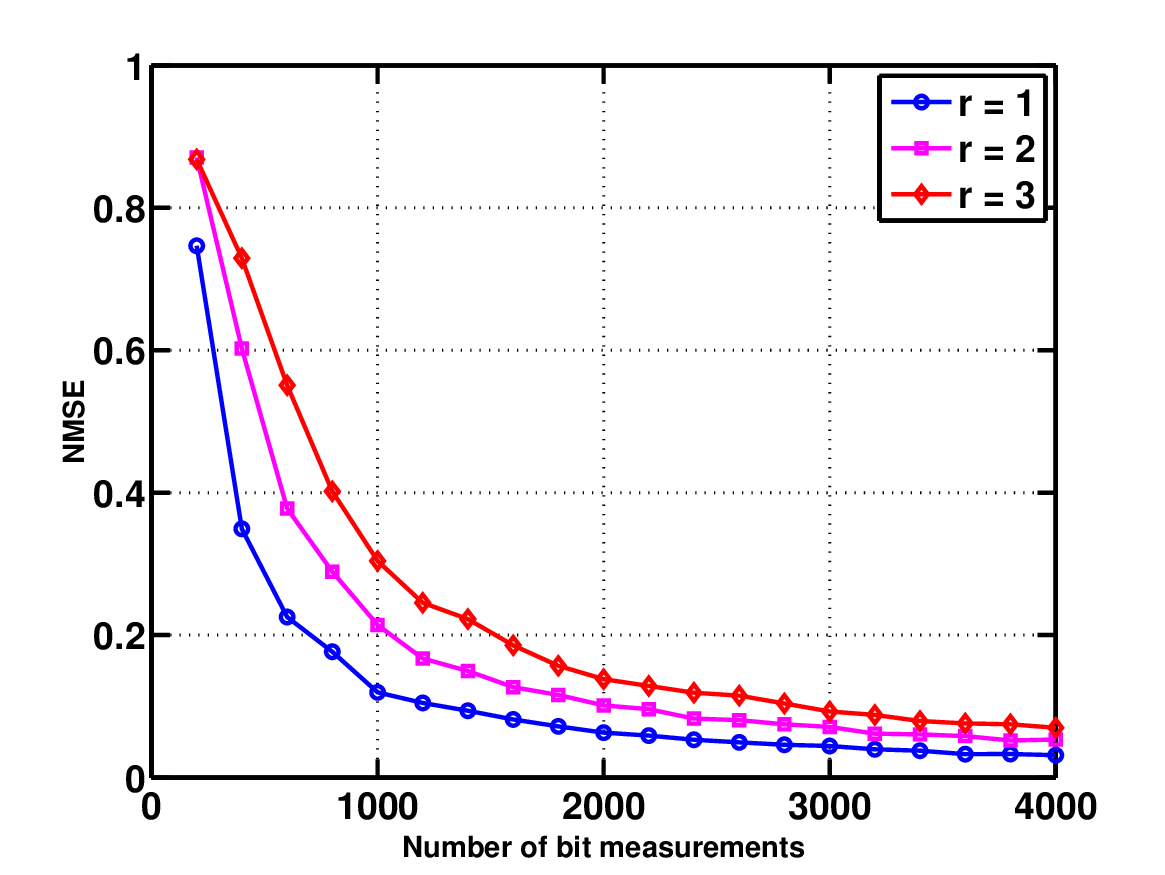} \\
(a)   & (b)  
\end{tabular}
\end{center}
\caption{Performance of the spectral estimator for estimating the principal subspace of low-rank covariance matrices. (a) NMSE with respect to the number of bit measurements for $n=40,100,200$ when $r=3$. (b) NMSE with respect to the number of bit measurements for $r=1,2,3$ when $n=100$.}\label{onebit_svd}
\end{figure*}

Let $\bR_m = (\bI-\bU_{m-1}\bU_{m-1}^H)\bK_m$ and $\bP_m=\mbox{orth}(\bR_m)$ be the orthonormal columns spanning the column space of $\bR_m$. We write $\bJ_m$ as
\begin{align*}
\bJ_m & = \begin{bmatrix}
\bU_{m-1} & \bP_m 
\end{bmatrix} \Bigg(\begin{bmatrix}
\eta_{m}\bPi_{m-1} &  0\\
0 & 0 
 \end{bmatrix} +  \\
& \quad  \begin{bmatrix}
\bU_{m-1}^H\bK_m \\
\bP_m^H\bR_m
 \end{bmatrix} \bLambda_m \begin{bmatrix}
\bU_{m-1}^H \bK_m \\
\bP_m^H\bR_m
 \end{bmatrix}^H  \Bigg)\begin{bmatrix}
\bU_{m-1}^H \\
 \bP_m^H
\end{bmatrix} \\
&: =  \begin{bmatrix}
\bU_{m-1} & \bP_m 
\end{bmatrix} \bGamma_{m} \begin{bmatrix}
\bU_{m-1}^H \\
 \bP_m^H 
\end{bmatrix},
\end{align*}
where $\bGamma_m$ is a small $(r+2)\times (r+2)$ matrix whose EVD can be computed easily and yields
$$ \bGamma_m  = \bU_{m}^{\prime}\bPi_m^{\prime} \bU_m^{\prime} .$$
Set $\bPi_m$ be the top $r\times r$ sub-matrix of $\bPi_m^{\prime}$ assuming the eigenvalues are given in an absolute descending order, the principal subspace of $\bJ_m$ can be updated correspondingly as
$$ \bU_m : =  \begin{bmatrix}
\bU_{m-1} & \bP_m 
\end{bmatrix} \bU_{m}^{\prime} \bI_{r},$$
where $\bI_r$ is the first $r$ columns of the $(r+2)\times (r+2)$ identity matrix.

\section{Numerical Experiments} \label{numerical}
In the numerical experiments, we first examine the performance of the spectral estimator in a batch setting in terms of reconstruction accuracy and robustness to flipping errors, with comparisons against the convex optimization algorithm in \eqref{algorithm}. We then examine the performance of the tailored spectral estimator when the covariance matrix is additionally Toeplitz. Next, we examine the performance of the online subspace estimation algorithm in Section~\ref{online} and apply it to the problem of line spectrum estimation. Finally, we examine the effects of aggregation over finite samples.

\subsection{Recovery for low-rank covariance matrices}\label{first_exp}
We generate the covariance matrix as $\bSigma = \bU\bU^T$, where $\bU\in\mathbb{R}^{n\times r}$ is composed of standard Gaussian entries. The one-bit measurements are then collected according to \eqref{noisy_1bit}. After the bit measurements are collected, we run the spectral estimator using the constructed $\bJ_m$ and the convex optimization algorithm \eqref{algorithm} assuming the rank $r$ of principal subspace is known perfectly. The algorithm \eqref{algorithm} is performed using the MOSEK toolbox available in CVX \cite{grant2008cvx} and obtains an estimate $\widehat{\bSigma}$, from which we extract its top-$r$ eigenvectors. The normalized mean squared error (NMSE) is defined as $\| (\bI-\widehat{\bU}\widehat{\bU}^H)\bU\|_{\F}^2/\|\bU\|_{\F}^2$, where $\widehat{\bU}$ is the estimated principal subspace with orthonormal columns.

\begin{figure}[htp]
\begin{center}
\includegraphics[width=0.4\textwidth]{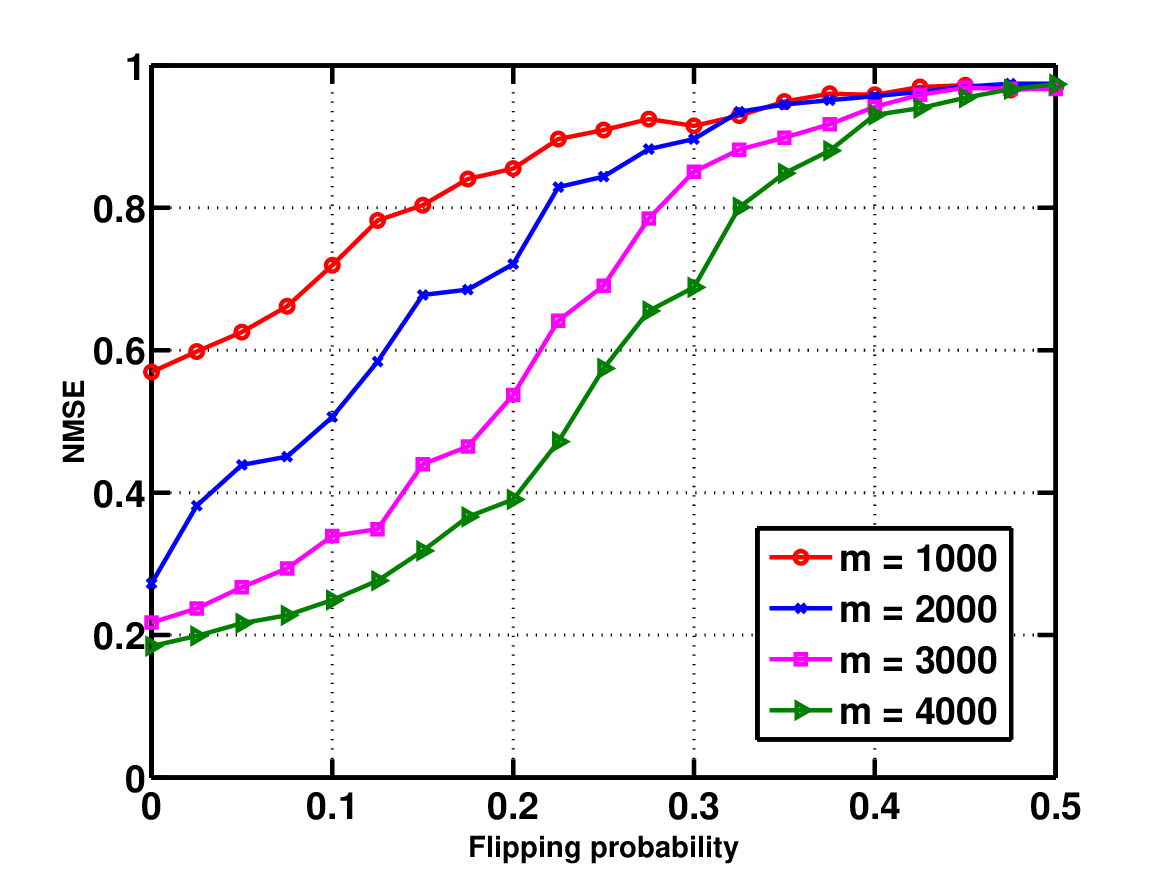} 
\caption{NMSE with respect to the flipping probability for different number of bit measurements when $n=100$ and $r=3$.}\label{var_flipping}
\end{center}
\end{figure}

Fig.~\ref{onebit_svd} (a) shows the NMSE with respect to the number of bit measurements for different $n=40,100, 200$ when $r=3$ averaged over $10$ Monte Carlo runs. Given the high complexity of the convex algorithm \eqref{algorithm}, we only perform it when $n=40$. We can see that its performance is comparable to that of the spectral estimator. Fig.~\ref{onebit_svd} (b) further examines the performance of the spectral estimator  for different ranks when $n=100$. For the same number of bit measurements, the NMSE grows gracefully as the rank increases. The spectral estimator also exhibits a reasonable robustness against flipping errors. Fig.~\ref{var_flipping} shows the reconstructed NMSE with respect to the flipping probability $p$ for different number of bit measurements $m$ when $n=100$ and $r=3$, where the error increases as the flipping probability $p$ increases.

\subsection{Recovery for low-rank Toeplitz covariance matrices}
We generate a rank-$r$ PSD Toeplitz matrix $\bSigma$ via its Vandermonde decomposition \cite{grenander1958toeplitz}, given as $\bSigma =\bV\bLambda \bV^{H}$, where $\bV =[\bv(\theta_1),\ldots, \bv(\theta_r)]\in \mathbb{C}^{n\times r}$ is a Vandermonde matrix with $\bv(\theta_k)=[1,e^{j\theta_k},\ldots,e^{j(n-1)\theta_k}]^T$, $\theta_k\in[0,1)$ for $1\leq k\leq r$, and $\bLambda = \mbox{diag}\left( [\sigma_1^2,\dots,\sigma_r^2 ] \right)$ is a diagonal matrix describing the power of each mode. 



\begin{figure}[htp]
\begin{center}
\includegraphics[width=0.4\textwidth]{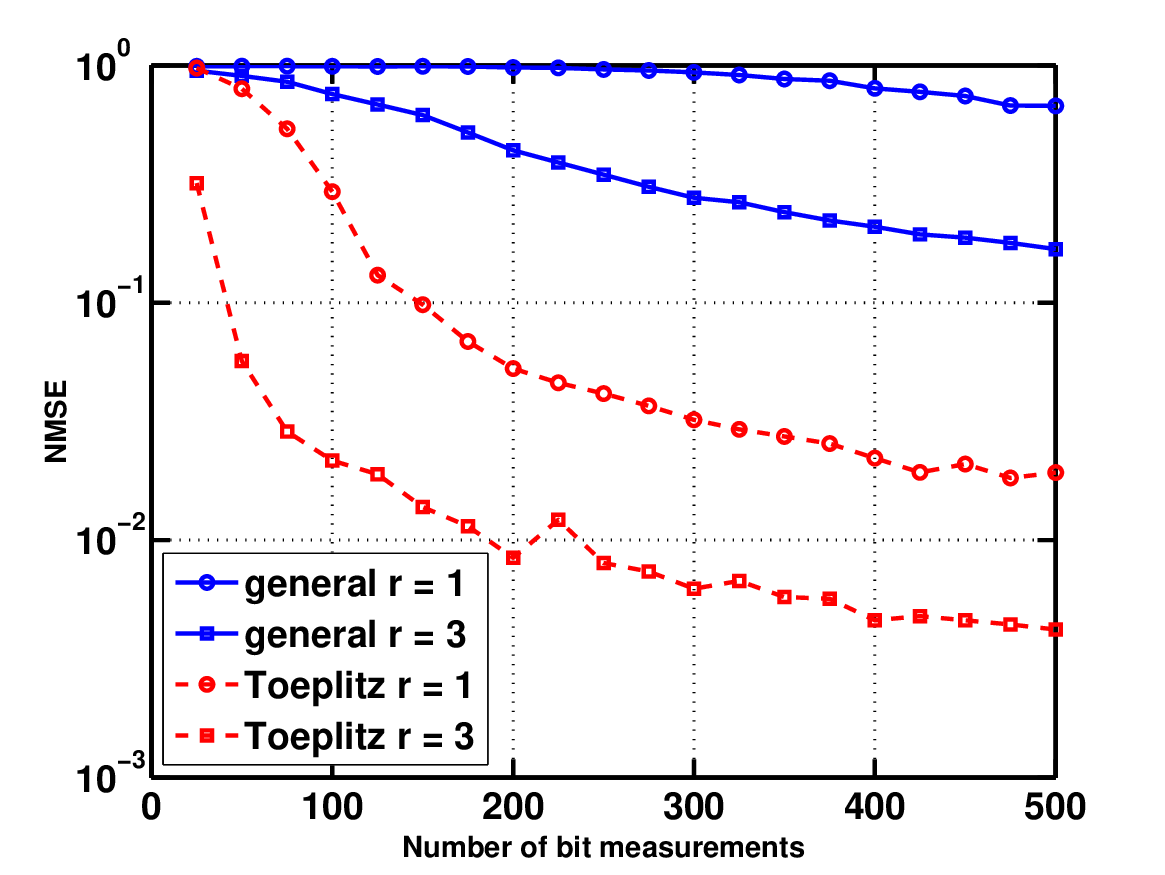} 
\end{center}
\caption{NMSE with respect to the number of bit measurements for estimating the principal subspace of low-rank Toeplitz covariance matrices when $n=40$ and $r=1$ and $r=3$.}\label{fig:rank_one}
\end{figure}

Fig.~\ref{fig:rank_one} depicts the NMSE with respect to the number of bit measurements for the spectral estimators using either $\bJ_m$ or $\cT(\bJ_m)$ when $n=40$, averaged over $100$ Monte Carlo simulations, when $r=1$ and $r=3$. It is clear that the tailored spectral estimator using $\cT(\bJ_m)$ achieves a smaller error using much fewer bits. Although Theorem~\ref{theorem_toeplitz} only applies to the rank-one case, the simulation suggests performance improvements even in the low-rank setting. 

 \begin{figure}[ht]
\begin{center}
\begin{tabular}{cc}
\hspace{-0.1in}\includegraphics[width=0.25\textwidth]{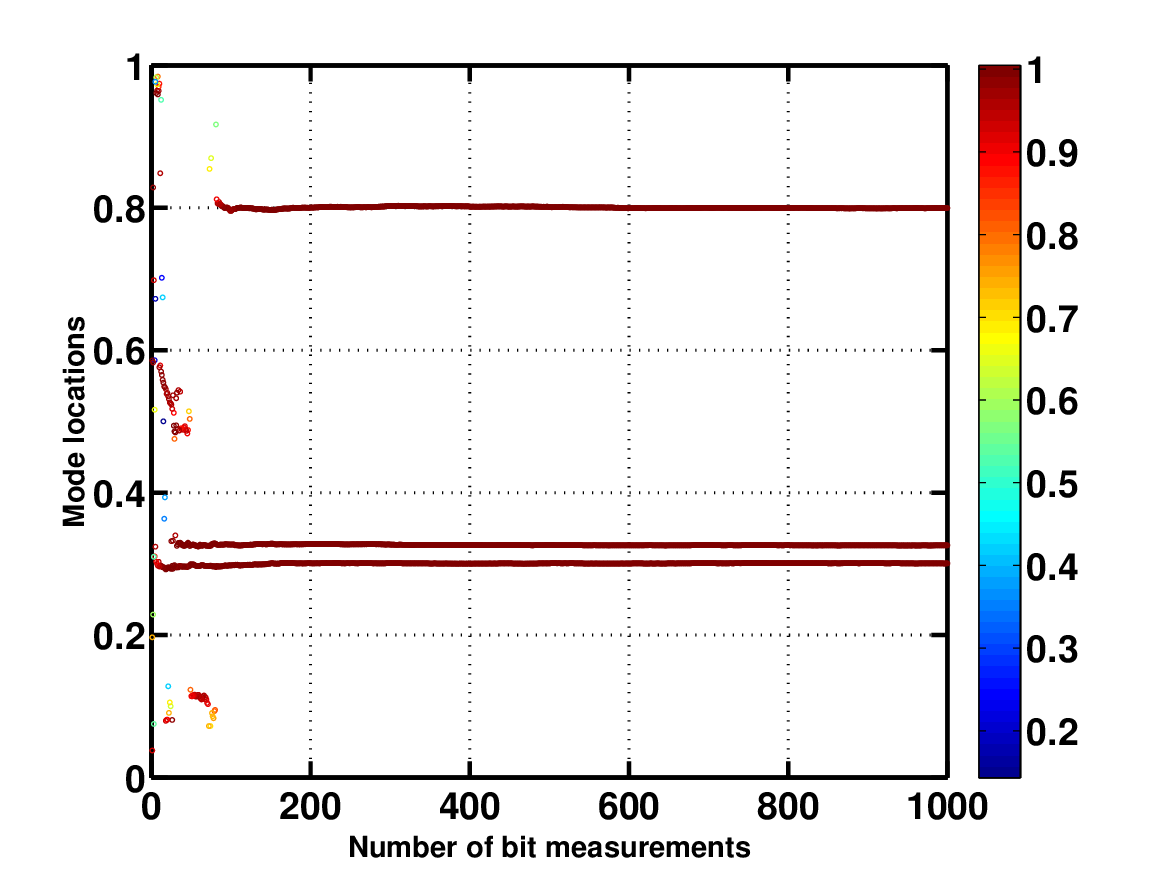} &\hspace{-0.2in}\includegraphics[width=0.25\textwidth]{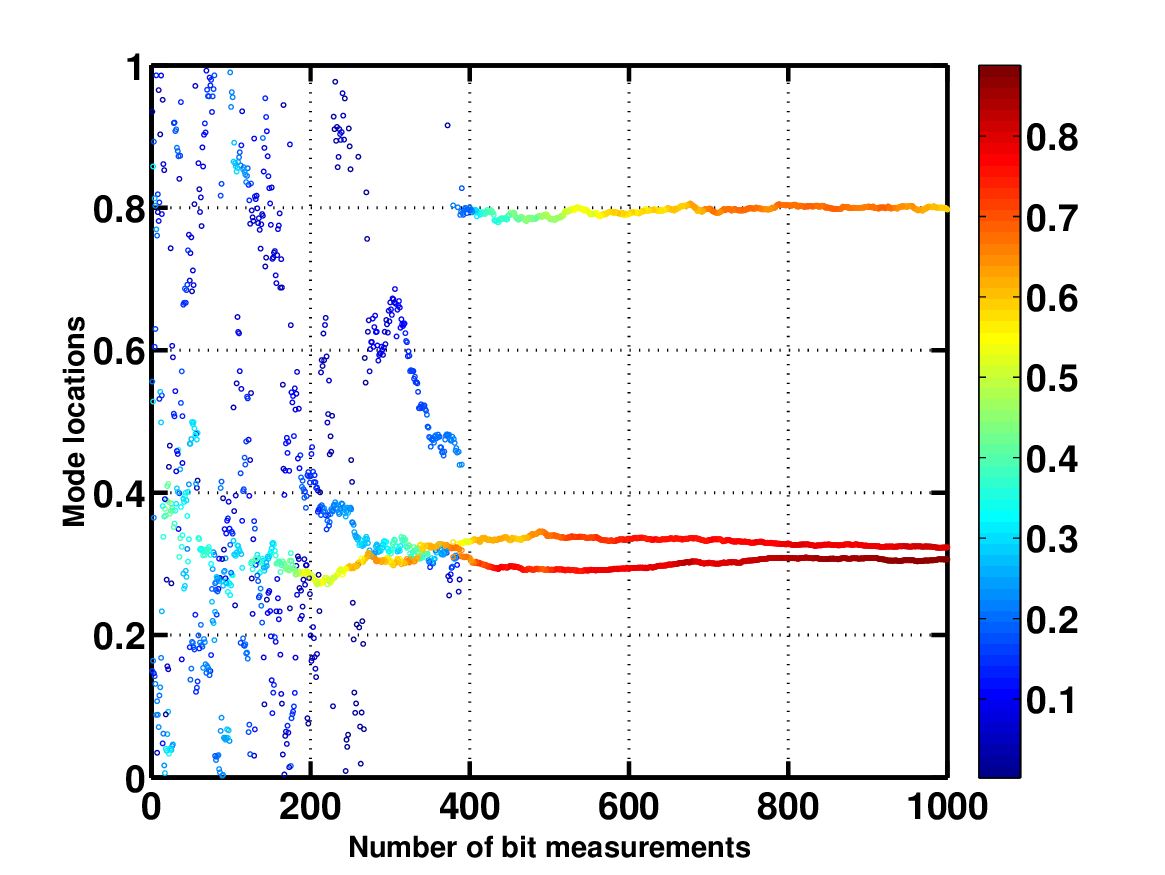}   \\
\hspace{-0.1in}(a)   & \hspace{-0.2in}(b)
\end{tabular}
\caption{Extracted mode locations using ESPRIT with respect to the number of bit measurements using the proposed spectral estimators using (a) $\cT(\bJ_m)$ and (b) $\bJ_m$, when $n=40$ and $r=3$. }\label{fig:low_rank}
\end{center}
\end{figure}
We perform ESPRIT \cite{RoyKailathESPIRIT1989} on the estimated subspace to recover the mode locations $\{\theta_k\}_{k=1}^r$. Fig.~\ref{fig:low_rank} shows the estimated mode locations vertically with respect to the number of bit measurements, with color indicating the power of the estimated modes, where the true mode locations are set as $[\theta_1,\theta_2,\theta_3] = [0.3, 0.325,0.8]$ and $[\sigma_1^2,\sigma_2^2,\sigma_3^2]=[1,1,0.5]$. The spectral estimator using $\cT(\bJ_m)$ estimates close-located modes and detects weak modes from a much smaller number of bits.

\subsection{Online subspace tracking}
\begin{figure}[htp]
\begin{center}
\includegraphics[width=0.4\textwidth]{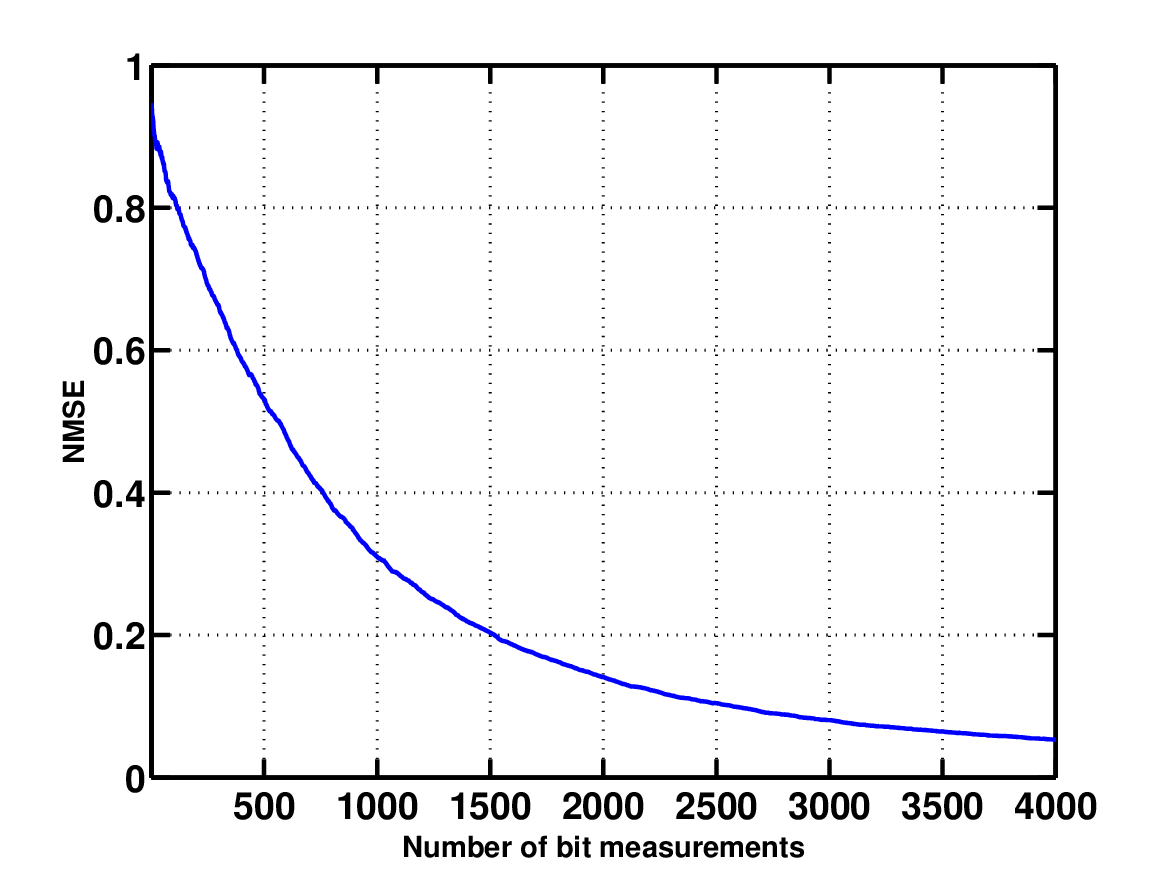}
\caption{The NMSE of principal subspace with respect to the number of bit measurements when $n=40$ and $r=3$ in an online setting. The estimation accuracy is comparable to that in a batch setting.}\label{onebit_online}
\end{center}
\end{figure}

We now examine the performance of the online subspace estimation algorithm proposed in Section~\ref{online}. Let $n=40$ and $r=3$.
Fig.~\ref{onebit_online} shows the NMSE of principal subspace estimation with respect to the number of bit measurements, where the result is averaged over $10$ Monte Carlo runs. Compared with Fig.~\ref{onebit_svd} (a), the estimation accuracy is comparable to that in a batch setting.

\begin{figure*}[htp]
\begin{center}
\begin{tabular}{ccc}
\hspace{-0.1in}\includegraphics[width=0.33\textwidth]{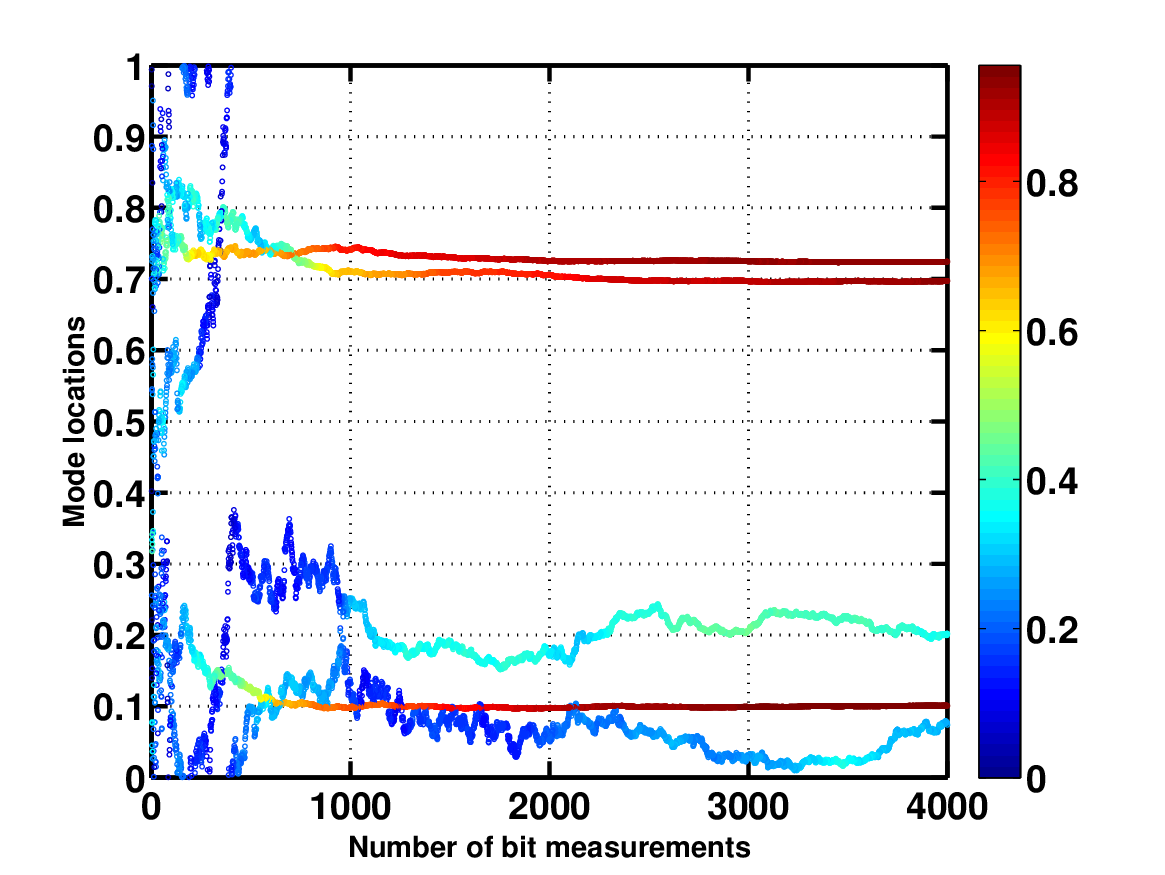} &\hspace{-0.1in}\includegraphics[width=0.33\textwidth]{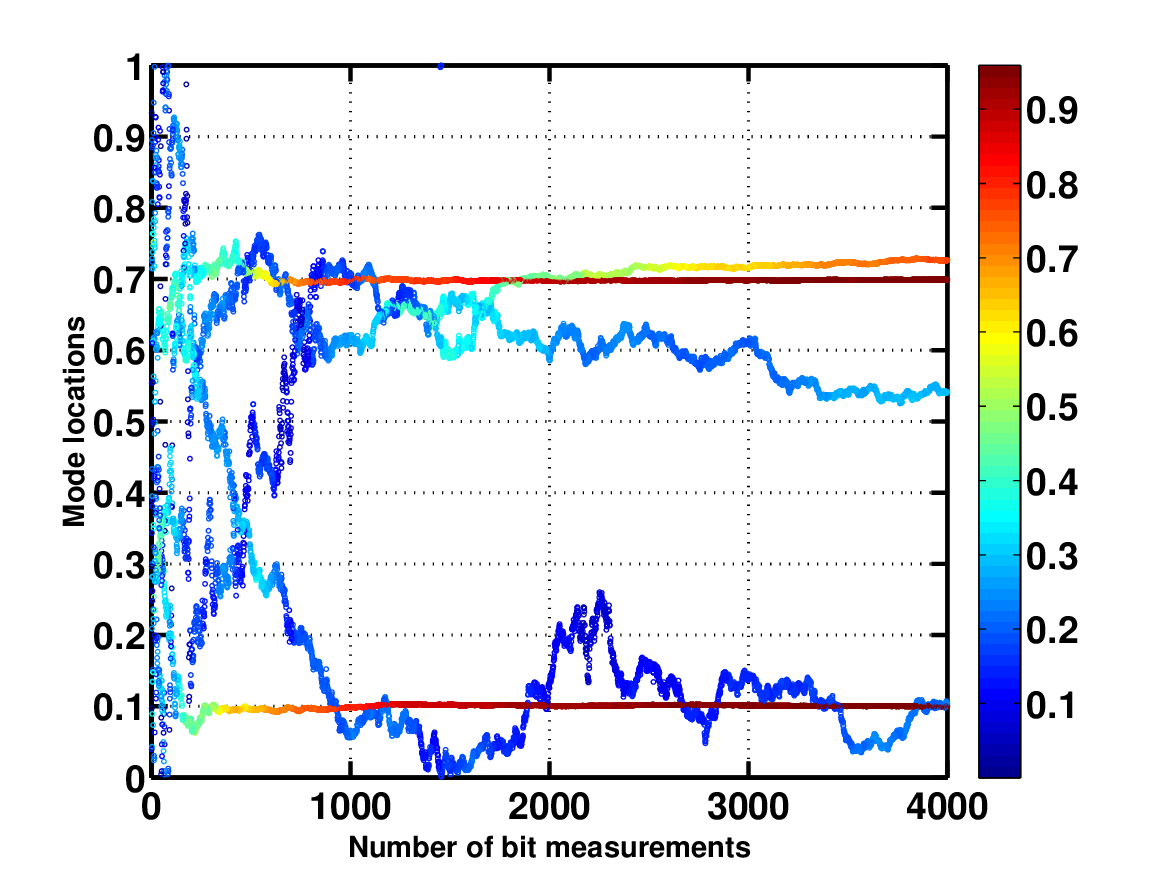} &
\hspace{-0.1in}\includegraphics[width=0.33\textwidth]{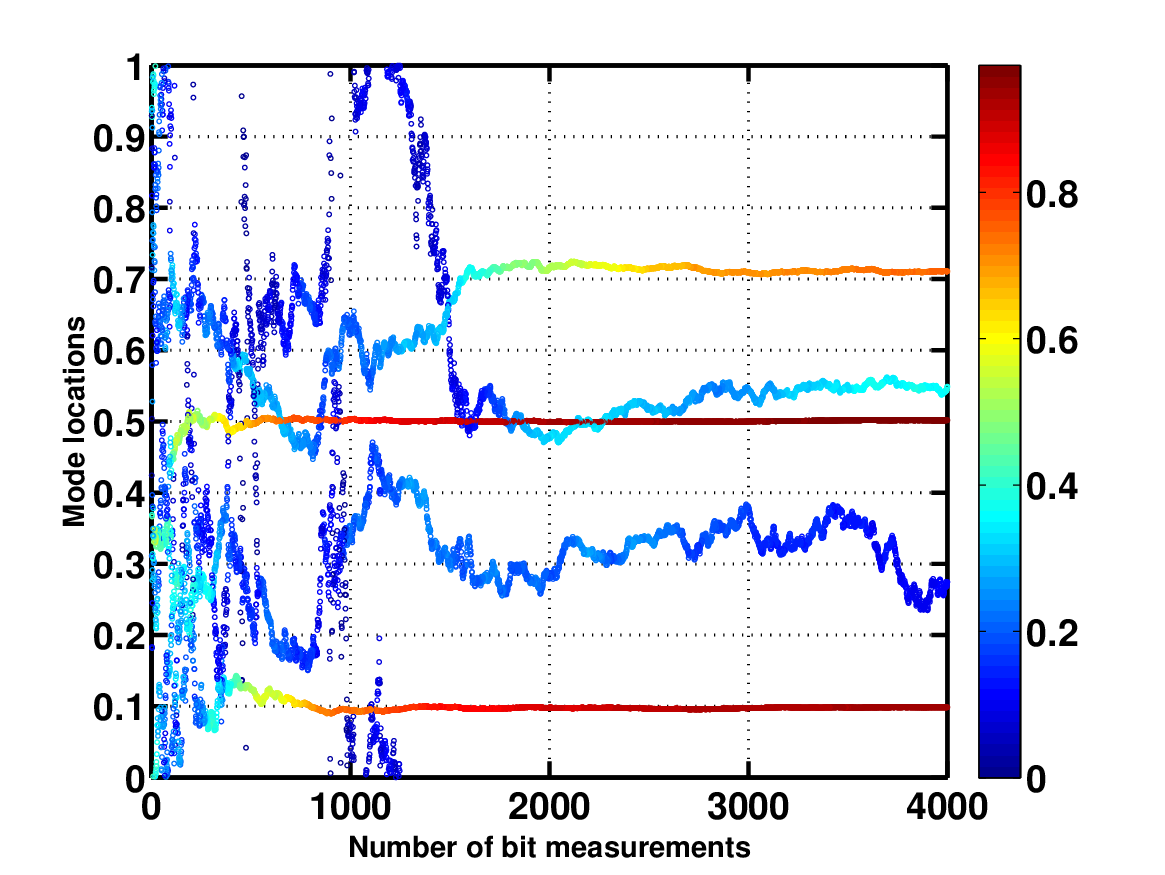} \\
\hspace{-0.1in} (a) $\begin{array}{c}
\cF   =[0.1,0.7,0.725] ; \\ 
\sigma^2  =[1, 1, 1];
\end{array}$ & \hspace{-0.1in}(b) $\begin{array}{c}
\cF   = [0.1,0.7,0.725] ; \\ 
\sigma^2  =[1, 1, 0.5];
\end{array}$ & \hspace{-0.1in} (c) $\begin{array}{c}
\cF   =[0.1,0.5,0.7] ; \\ 
\sigma^2  =[1, 1, 0.3];
\end{array}$
\end{tabular}
\end{center}
\caption{Online line spectrum estimation: the estimated frequency values against the number of bit measurements when $n=40$ and $r=3$. Each time $5$ frequencies are estimated with the color bar indicates their amplitudes. The true frequency profiles and their amplitudes are given in the subtitles.}\label{onebit_doa}
\end{figure*}


In the sequel, we apply the online estimator to the problem of line spectrum estimation, in a set up similar to Fig.~\ref{fig:low_rank}. Note here we do not exploit the additional Toeplitz structure of $\bSigma$. At each new binary measurement, we first use the online subspace estimation algorithm proposed in Section~\ref{online} to estimate a principal subspace of rank ${r}_{\text{est}}=5$, then apply ESPRIT \cite{RoyKailathESPIRIT1989} to recover the mode locations. Fig.~\ref{onebit_doa} shows the estimation results for various parameter settings, where the estimates of mode locations are plotted vertically at each new bit measurement. Fig.~\ref{onebit_doa} (a) and (b) have the same set of modes, with two close located frequencies separated by the Rayleigh limit, $1/n$. When all the modes have strong powers, the modes can be accurately estimated as depicted in (a); when one of the close modes is relatively weak, the algorithm requires more measurements to pick up the weak mode, as depicted in (b). Fig.~\ref{onebit_doa} (c) examines the case when all the modes are well separated, and one of them is weak. The algorithm picks up a weak mode with a smaller number of measurements when the modes are well separated. Taking these together, it suggests that the tracking performance depends on the eigengap and the conditioning number of the covariance matrix. 

\subsection{Performance with finite data samples}

The above simulations assume that the bit measurements are exact. We now examine the performance of the spectral estimator assuming it is measured via \eqref{sample_sign} using a finite number of data samples. Let $n=100$ and $r=3$. Assume there are a collection of $T$ samples generated as $\bx_t = \bU\ba_t + \bn_t$, where $\bU$ is an orthogonal matrix normalized from a random matrix generated with i.i.d. Gaussian entries, $\ba_t\sim\mathcal{N}(0,\bI_r)$ is generated with standard Gaussian entries, and $\bn_t\sim\mathcal{N}(0,\sigma^2\bI_n)$ is independently generated Gaussian entries. In other words, $\bx_t\sim\mathcal{N}(\bU\bU^T+\sigma^2\bI_n)$. All $m$ sensors measure the same set of $T$ samples (i.e. $\ell_t^i=i$, for $i=1,\cdots, T$) and communicate their bit measurements for subspace estimation. 

\begin{figure*}[t]
\begin{center}
\begin{tabular}{cc}
\includegraphics[width=0.4\textwidth]{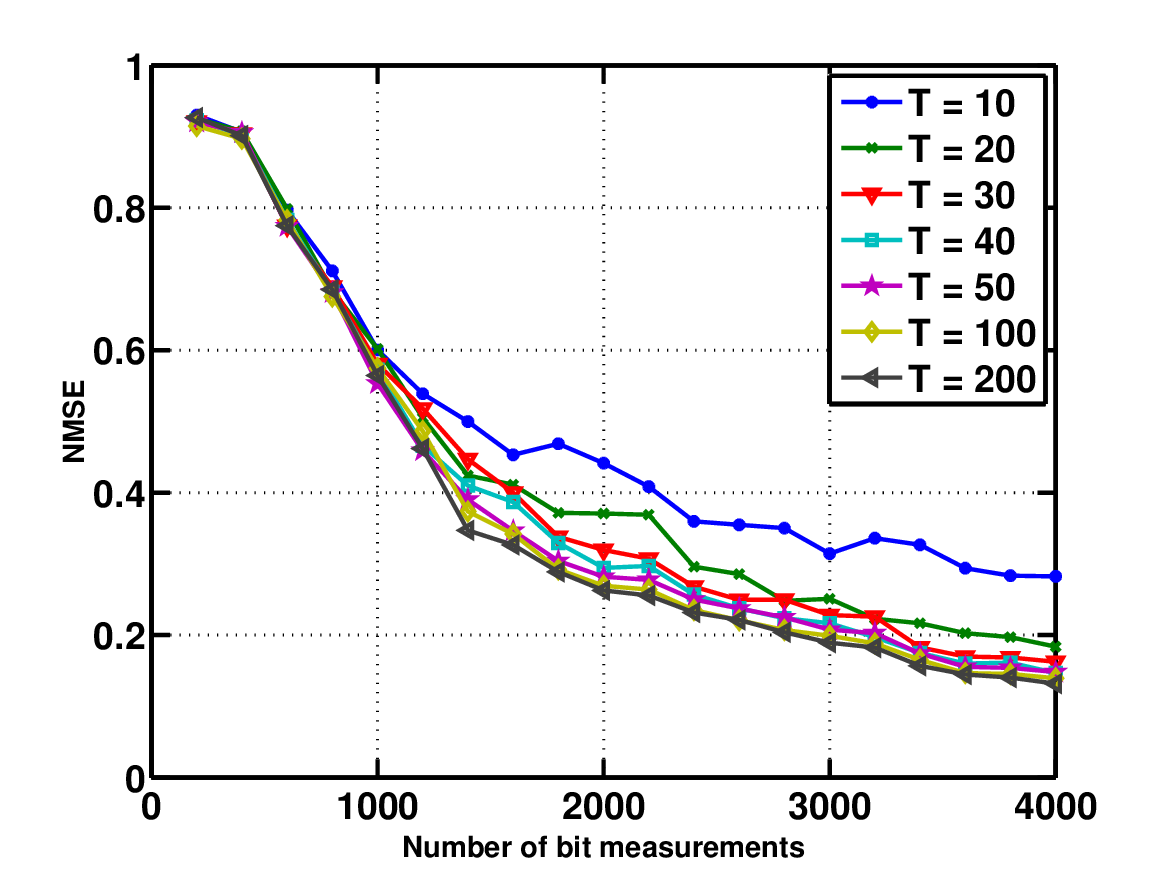} & \includegraphics[width=0.4\textwidth]{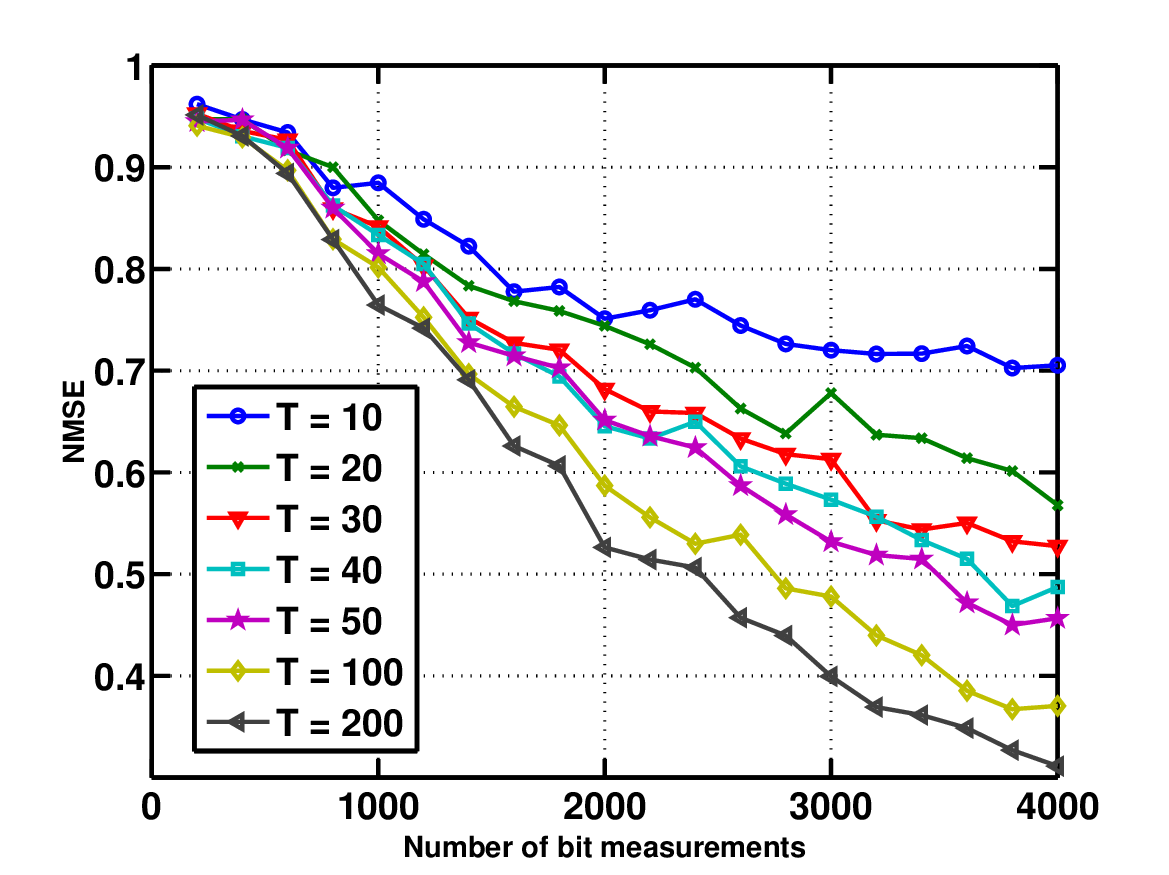} \\
(a) $\sigma^2= 0$ &  (b)  $\sigma^2= 0.1$
\end{tabular}
\caption{The NMSE of principal subspace estimate with respect to the number of bit measurements when $n=40$ and $r=3$ for different number of samples for (a) noise-free samples and (b) noisy samples.}\label{finite_sample}
\end{center}
\end{figure*}

Fig.~\ref{finite_sample} shows the NMSE of principal subspace estimate with respect to the number of bit measurements for different number of samples $T=10,20,30,40,50,100$ and $200$ averaged over $20$ Monte Carlo runs when the samples are (a) noise-free with $\sigma^2= 0$, and (b) noisy with $\sigma^2=0.1$. As $T$ increases, the NMSE decreases as the bit measurements get more accurate in light of Theorem~\ref{all_bits_sample}. Note that the gain diminishes as $T$ is sufficiently large as all bit measurements are accurate with high probability. For noisy data samples, it is evident that more samples are necessary for the aggregation procedure to yield accurate bit measurements, and performance improves as more samples are averaged. 





%

\section{Conclusions}\label{conclusion}
In this paper, we present a low-complexity distributed sensing and central estimation framework to recover the principal subspace of low-rank covariance matrices from a small number of one-bit measurements based on aggregated energy comparisons of the data samples. Spectral estimators are proposed with appealing computational complexity and theoretical performance guarantees. In the future, it is of interest to develop principal subspace estimation algorithms from quantized measurements beyond the one-bit scheme exploited in this paper.  
 
\section*{Acknowledgement}
The authors thank useful feedbacks from the anonymous reviewers that significantly improve the quality of this paper. The first author thanks Lee Seversky, Lauren Huie and Matt Berger for their hospitality and helpful discussions during her stay at the Air Force Research Lab, Rome, New York where part of this work was accomplished. She also thanks Yuxin Chen for helpful discussions. 

\appendices

\section{Supporting Lemmas}

\begin{lemma}[scalar Bernstein's inequality with sub-exponential norm, \cite{van1996weak}]\label{scalar_bernstein}
Let $z_1, \ldots, z_L $ be independent random variables with $\mathbb{E}[z_k]=0$ and $\sigma_k^2 = \mathbb{E}[z_k^2]$, and $\mathbb{P}[|z_k|>u ]\leq Ce^{-u/\sigma_k}$ for some constants $C$ and $\sigma_k$. Define $\sigma^2 = \sum_{k=1}^L \sigma_k^2$ and $B = \max_{1\leq k\leq L} \sigma_k$. Then
$$ \mathbb{P}\left[ \left|\sum_{k=1}^L z_k \right| > u \right] \leq 2 \exp\left(- \frac{u^2}{2C\sigma^2 + 2B u} \right). $$
\end{lemma} 

\begin{lemma}[matrix Bernstein's inequality with sub-exponential norm, \cite{tropp2012user}]\label{matrix_bernstein}
Let $\bX_1$, $\cdots$, $\bX_L$ be independent zero-mean symmetric random matrices of dimension $n \times n$. Suppose $\sigma^2= \left\|\sum_{k=1}^L\mathbb{E}[ \bX_k\bX_k^{H}]\right\|$ and $\left\|\bX_k\right\|_{\psi_1}\leq B$ almost surely for all $k$, where $\|\cdot\|_{\psi_1}$ is the Orlitz norm \cite{Vershynin2012}. Then for any $\tau>0$,
\begin{align}
\mathbb{P}\left[ \left\|\sum_{k=1}^L \bX_k \right\| > \tau \right] & \leq 2n \exp\left( -\frac{\tau^2}{2\sigma^2 + 2B\tau/3 }\right) .
\end{align}
\end{lemma} 

\begin{lemma}[Davis-Kahan, \cite{lei2013consistency}]\label{davis_kahan}
Let $\bA, \tilde{\bA}\in\mathbb{R}^{n\times n}$ be symmetric matrices. Denote $\bV$ as the subspace spanned by the top $r$ eigenvalues of $\bA$, and $\tilde{\bV}$ as the subspace spanned by the top $r$ eigenvalues of $\tilde{\bA}$. Let $\delta$ be the spectral gap between the $r$th and the $(r+1)$th eigenvalue of $\bA$. Then there exists an $r\times r$ orthogonal matrix $\bQ$ such that the two subspaces $\bV$ and $\tilde{\bV}$ is bounded by
$$\| \tilde{\bV} - \bV \bQ \|_{\mathrm{F}}  \leq \max\left\{ \frac{ \sqrt{2} \|  \tilde{\bA} -\bA \|_{\F} }{\delta}, \frac{2\sqrt{2r}\|  \tilde{\bA} -\bA \| }{\delta} \right\}. $$
\end{lemma}

\begin{lemma}[Hanson-Wright inequality, \cite{rudelson2013hanson}] \label{hanson_wright}
Let $\bSigma$ be a fixed $n\times n$ matrix. Consider a random vector $\bx=(X_1, \ldots, X_n)$ where $X_i$ are independent random variables satisfying $\mathbb{E}X_i = 0$ and $\|X_i\|_{\psi_2} \leq K$. Then for any $t \geq 0$, we have
\begin{equation*}
\mathbb{P}\left[\left| \bx^H\bSigma \bx  - \mathbb{E}\bx^H\bSigma \bx  \right| >t \right]  \leq 2e^{  -c \min \left(\frac{t^2}{K^4 \| \bSigma \|_{\mathrm{F}}^2}, \frac{t}{K^2 \|\bSigma\|} \right) } .
\end{equation*}
\end{lemma}

\section{Proof of Proposition~\ref{single_bit_sample}} \label{proof_single_bit_sample}
\begin{proof} For notational simplicity, we drop the sensor index $i$ and let $\bSigma_T = \frac{1}{T} \sum_{i=1}^T \bx_i\bx_i^H$ and $y_{i,T} = \langle \bW_i, \bSigma_T \rangle $. Conditioned on $\ba_i$ and $\bb_i$, we have $\mathbb{E}[y_{i,T} |\ba_i,\bb_i ] = \langle \bW_i, \bSigma \rangle$. Let
$$ y_{i,T} -  \mathbb{E}[y_{i,T} |\ba_i, \bb_i ] = \frac{1}{T} \sum_{t=1}^T  \langle \bW_i, \bSigma - \bx_t\bx_t^H \rangle := \frac{1}{T} \sum_{t=1}^T Q_t, $$
where $Q_t= \langle \bW_i, \bSigma - \bx_t\bx_t^H \rangle $. We may appeal to the Bernstein-type inequality in Lemma~\ref{matrix_bernstein}. First, $\mathbb{E}[Q_t | \ba_i, \bb_i]=0$. Second,
\begin{align*}
\Var[Q_t | \ba_i, \bb_i] & = \Var[\langle \bW_i, \bx_t\bx_t^H \rangle | \ba_i, \bb_i ] \\
& = \Var \left[ |\ba_i^H \bx_t|^2 - |\bb_i^H \bx_t|^2 \big| \ba_i,\bb_i \right] \\
& \leq 2 \mathbb{E} [|\ba_i^H \bx_t|^2 | \ba_i] + 2 \mathbb{E} [|\bb_i^H \bx_t|^2| \bb_i] \\
&  = 2(\ba_i^H\bSigma \ba_i + \bb_i^H \bSigma \bb_i): = 2B,
\end{align*}
where $\bx_t^H\ba_i| \ba_i \sim \mathcal{CN}(0, \ba_i^H \bSigma \ba_i)$ and $\bx_t^H\bb_i| \bb_i \sim \mathcal{CN}(0, \bb_i^H \bSigma \bb_i)$ are two correlated Gaussian variables. 
Third, 
\begin{align*}
 | Q_t | & = |  \langle \bW_i, \bSigma - \bx_t\bx_t^H \rangle | \\
 & \leq  \left| \ba_i^H\bSigma \ba_i - \bb_i^H \bSigma \bb_i - | \bx_t^H \ba_i |^2 + |\bx_t^H\bb_i|^2 \right| \\
& \leq \ba_i^H \bSigma \ba_i + \bb_i^H \bSigma \bb_i + |\bx_t^H \ba_i|^2 +  |\bx_t^H \bb_i|^2,
\end{align*}
Since $ |\bx_t^H \ba_i|^2/\ba_i^H\bSigma \ba_i$ and $|\bx_t^H \bb_i|^2/\bb_i^H \bSigma \bb_i$ are exponential random variables with parameter $1$,  we have
$$ \|Q_t\|_{\psi_1} \leq 2(\ba_i^H \bSigma \ba_i + \bb_i^H \bSigma \bb_i):= 2B. $$

Assume $\langle \bW_i, \bSigma \rangle $ is positive without loss of generality, we have
\begin{equation}\label{conditional}
\begin{split}
&  \mathbb{P}\left[|y_{i,T} -  \mathbb{E}[y_{i,T}|\ba_i, \bb_i ] | > \langle \bW_i, \bSigma \rangle | \ba_i, \bb_i \right] \ \\
 & \leq 2\exp \left( - \frac{|\langle \bW_i, \bSigma \rangle |^2 T}{4B (1+|\langle \bW_i,\bSigma \rangle| /3)} \right)
 \end{split}
 \end{equation}

Next, we can bound the quadratic form $\ba_i^H\bSigma \ba_i$ using the Hanson-Wright inequality \cite{rudelson2013hanson} in Lemma~\ref{hanson_wright}. Since $\mathbb{E}[\ba_i^H\bSigma\ba_i] = \mbox{Tr}(\bSigma)$, with probability at least  $1-\delta/3$, we have
$$ \left|\ba_i^H\bSigma\ba_i - \mbox{Tr}(\bSigma)\right| \leq c \|\bSigma\|_{\mathrm{F}} \log(1/\delta) $$
for some constant $c$. Since $\|\bSigma\|_{\mathrm{F}} \leq \mbox{Tr}(\bSigma)$, we have $\ba_i^H\bSigma \ba_i \leq  c_1\mbox{Tr}(\bSigma)\log(1/\delta)$ with probability at least  $1-\delta/3$ for some absolute constant $c_1$. Denote this as event $\mathcal{G}_1$. Further from the arguments in \cite[Proposition 1]{chen2013exact}, we have that 
\begin{equation}\label{bound_inner}
c_2 \|\bSigma\|_{\mathrm{F}}\log(1/\delta) \leq |\langle \bW_i,\bSigma \rangle | \leq c_3 \|\bSigma\|_{\mathrm{F}}\log(1/\delta)
\end{equation}
with probability at least $1-\delta/3$ for some absolute constants $c_2$ and $c_3$. Denote this as event $\mathcal{G}_2$. 

Conditioned on the event $\mathcal{G}_1$ and $\mathcal{G}_2$, and plug in the above into \eqref{conditional}, we have as soon as 
\begin{equation}\label{sample_complexity} 
T\geq c_4 \frac{\mathrm{Tr}(\bSigma)}{\|\bSigma\|_{\mathrm{F}}}\log^2 (1/\delta) 
\end{equation}
for some constant $c_4$, the RHS of \eqref{conditional} can be upper bounded by $\delta/3$. To summarize, assuming \eqref{sample_complexity} holds,
\begin{align*}
& \mathbb{P}\left[ y_i \neq  y_{i,T}  \right] \leq \int  \mathbb{P}\left[  y_i \neq  y_{i,T}   | \ba_i, \bb_i \right]  d\mu(\ba_i) d\mu(\bb_i) \\
& \leq \mathbb{P}(\mathcal{G}_1^c) +  \mathbb{P}(\mathcal{G}_2^c) + \\
&   \int_{\mathcal{G}_1,\mathcal{G}_2} \mathbb{P}\left[|y_{i,T} -  \mathbb{E}[y_{i,T}|\ba_i, \bb_i ] | >  \langle \bW_i, \bSigma \rangle | \ba_i, \bb_i \right] d\mu(\ba_i) d\mu(\bb_i)\\
&   \leq \delta.
\end{align*}
Our proposition then follows.
\end{proof}
%

\section{Approximate Low-Rank Covariance Matrices}\label{approx_lr}

Without loss of generality, assume $\langle \bW_i,\bSigma \rangle$ is positive. We wish $\langle \bW_i,\bSigma_r \rangle$ is also positive so that they have the same sign. Note that $ \langle \bW_i,\bSigma_r \rangle  = \langle \bW_i,\bSigma \rangle - \langle \bW_i,\bSigma - \bSigma_r \rangle$, it is sufficient to have
\begin{equation}\label{sign_same}
\langle \bW_i,\bSigma - \bSigma_r \rangle \leq   |\langle \bW_i,\bSigma \rangle| 
\end{equation}
for all $i=1,\ldots, m$. Since
\begin{align*}
|\langle \bW_i,\bSigma - \bSigma_r \rangle| &  \leq \| \bW_i\| \cdot \| \bSigma - \bSigma_r \|_* \\
& \leq (\| \ba_i\|_2^2 + \| \bb_i\|_2^2  )\cdot \| \bSigma - \bSigma_r \|_* ,
\end{align*}
and from \cite{wainwright2009information}, with probability at least $1-4\delta$, we have
$$\max_i \{ \| \ba_i\|_2^2, \| \bb_i\|_2^2\} \leq  n (1+2\sqrt{\log(m/\delta)}). $$
Combined with \eqref{bound_inner}, and renaming the constants, then \eqref{sign_same} is guaranteed with probability at least $1-\delta$, as long as
$$  \| \bSigma - \bSigma_r \|_*   \leq \frac{c\|\bSigma\|_\F\log(m/\delta)}{n (1+2\sqrt{\log(m/\delta)})}$$
for some constant $c$.
 
\section{Proof of Lemma~\ref{evd_expectedJ}}\label{proof_evd_expectedJ}

\begin{proof} 
We first prove \eqref{minor}. For $k=1,\ldots, n-r$,
\small
\begin{align*}
\frac{\bv_k^H {\bJ} \bv_k}{(1-2p)} 
& = \mathbb{E}\Bigg[\frac{1}{m}\sum_{i=1}^m \mbox{sign}\left(\sum_{k=1}^r \lambda_k |\ba_i^H\bu_k|^2 -\sum_{k=1}^r \lambda_k |\bb_i^H\bu_k|^2  \right) \\
& \quad \quad \cdot\left(|\ba_i^H \bv_k|^2 - |\bb_i^H \bv_k|^2 \right)\Bigg] \\
& = \mathbb{E}\left[  \mbox{sign}\Bigg(\sum_{k=1}^r \lambda_k |\ba_i^H\bu_k|^2 -\sum_{k=1}^r \lambda_k |\bb_i^H\bu_k|^2  \right) \\
& \quad\quad \cdot \left(|\ba_i^H \bv_k|^2 - |\bb_i^H \bv_k|^2 \right)\Bigg], \\
& = \mathbb{E}\left[  \mbox{sign}\left(\sum_{k=1}^r \lambda_k |\ba_i^H\bu_k|^2 -\sum_{k=1}^r \lambda_k |\bb_i^H\bu_k|^2  \right) \right] \\
& \quad\quad \cdot \mathbb{E}\left[|\ba_i^H \bv_k|^2 -  |\bb_i^H \bv_k|^2 \right] =0,
\end{align*}
\normalsize
where the penultimate equation follows from that $\ba_i^H\bv_k$'s and $\ba_i^H \bu_k$'s are independent from the Gaussianity of $\ba_i$, and the last equality follows from $\mathbb{E}\left[|\ba_i^H \bv_k|^2 - |\bb_i^H \bv_k|^2 \right]=0$. 

We next prove \eqref{principal}. For $k=1,\ldots, r$, 
\small
\begin{align*}
 \frac{\bu_k^H\bJ\bu_k}{(1-2p)} &  = \mathbb{E}\Bigg[  \mbox{sign}\left(\sum_{k=1}^r \lambda_k |\ba_i^H\bu_k|^2 -\sum_{k=1}^r \lambda_k |\bb_i^H\bu_k|^2  \right) \\
& \quad\quad \cdot \left(|\ba_i^H \bu_k|^2 - |\bb_i^H \bu_k|^2 \right)\Bigg] \\
&  = \mathbb{E}\left[  \mbox{sign}\left(\sum_{k=1}^r \lambda_k V_k  \right)V_k\right],
\end{align*}
\normalsize
where $V_k= |\ba_i^H \bu_k|^2 - |\bb_i^H \bu_k|^2$'s are i.i.d. random variables following the Laplace distribution with parameter $1$. Let $E = \sum_{k'\neq k} \lambda_{k'} V_{k'} $, then by iteratively applying conditional expectation,
\begin{align*}
\mathbb{E}\left[  \mbox{sign}\left(\sum_{k=1}^r \lambda_k V_k  \right)V_k\right]&   = \mathbb{E}\left[ \mathbb{E}\left[  \mbox{sign}\left(\lambda_kV_k + E \right)V_k| E=\epsilon \right]\right]\\
& \hspace{-0.1in}=\mathbb{E}\left[ \mathbb{E}\left[  \mbox{sign}\left( V_k + \epsilon/\lambda_k \right)V_k| E=\epsilon \right]\right] .
\end{align*}
First, assume $\epsilon>0$,
\begin{align*}
 &\quad \mathbb{E}\left[  \mbox{sign}\left( V_k +  \epsilon/\lambda_k \right)V_k| E  =\epsilon \right] \\
 & = \int_{-\infty}^{-\epsilon/\lambda_k} (-v)f_{V_k}(v) dv + \int_{-\epsilon/\lambda_k}^{\infty}  v f_{V_k}(v) dv  \\
 & = 2\int_{\epsilon/\lambda_k}^{\infty} vf_{V_k}(v) dv  = (1+\epsilon/\lambda_k)e^{-\epsilon/\lambda_k}.
\end{align*}
Similarly we derive it for $\epsilon<0$, together we have
$$  \mathbb{E}\left[  \mbox{sign}\left( V_k +  \epsilon/\lambda_k \right)V_k| E  =\epsilon \right]  = (1+|\epsilon| /\lambda_k)e^{-|\epsilon|/\lambda_k}. $$
It is straightforward that $(1+|\epsilon| /\lambda_k)e^{-|\epsilon|/\lambda_k}\leq 1$, therefore $\bu_k^H {\bJ}\bu_k\leq  (1-2p)$. On the other hand,
\begin{align}
\frac{\bu_k^H {\bJ}\bu_k}{(1-2p) } & = \mathbb{E}\left[ (1+|E| /\lambda_k)e^{-|E|/\lambda_k} \right] \nonumber \\
& \geq \mathbb{E} [e^{-|E|/\lambda_k}] = \mathbb{E} \left[e^{- \left| \sum_{k'\neq k} \lambda_{k'} V_{k'} \right|/\lambda_k} \right]. \label{proof_six}
\end{align}
Next, we provide two lower bounds on \eqref{proof_six}, then \eqref{principal} follows by taking the maximum of the two bounds. The first lower bound follows straightforwardly from
\begin{align}
 \mathbb{E} \left[e^{- \left| \sum_{k'\neq k} \lambda_{k'} V_{k'} \right|/\lambda_k} \right]  &\geq \mathbb{E} \left[e^{-\sum_{k'\neq k} \lambda_{k'} \left|  V_{k'} \right|/\lambda_k} \right]  \nonumber\\ 
 &= \prod_{k'\neq k} \mathbb{E}\left[e^{-\lambda_{k'}|V_{k'}|/\lambda_k}\right] \label{independence_Vk} \\
& =  \prod_{k'\neq k} 2 \int_0^{\infty} e^{-\lambda_{k'}v/\lambda_k} f_{V_{k'}}(v) dv \nonumber \\
& = \prod_{k'\neq k}  \frac{\lambda_{k}}{\lambda_{k}+\lambda_{k'}},\label{prodexp}
\end{align}
where \eqref{independence_Vk} follows from the independence of $V_k$'s. Combining \eqref{proof_six} and \eqref{prodexp}, we obtain
\begin{align*} 
\frac{\bu_k^H \bJ \bu_k}{(1-2p) } & \geq   \prod_{k'\neq k} \frac{\lambda_{k}}{\lambda_{k}+\lambda_{k'}}  \geq \left(\frac{1}{1+\kappa(\bSigma)}\right)^{r-1}.
\end{align*}
The second lower bound follows from
\begin{align*}
 &\quad\; \;\mathbb{E} \left[e^{- \left| \sum_{k'\neq k} \lambda_{k'} V_{k'} \right|/\lambda_k} \right] \\
 & = \int_{0}^{\infty} \mathbb{P}\left[e^{- \left| \sum_{k'\neq k} \lambda_{k'} V_{k'} \right|/\lambda_k} \geq h \right] dh \\
 & = \int_{0}^{\infty} \mathbb{P}\left[ \left| \sum_{k'\neq k} \frac{\lambda_{k'}}{\lambda_k} V_{k'} \right| \leq  \log\left(\frac{1}{h}\right) \right] dh  \\
 & \geq \mathbb{P}\left[ \left| \sum_{k'\neq k} \frac{\lambda_{k'}}{\lambda_k} V_{k'} \right| \leq \kappa(\bSigma) \right]  e^{-\kappa(\bSigma)} 
\end{align*}
where the last equation is obtained by setting $h=e^{-\kappa(\bSigma)}$. Applying Lemma~\ref{scalar_bernstein} with $u=\kappa(\bSigma)$, $B = \sqrt{2}\kappa(\bSigma)$, $C=1$ and $\sigma^2 =2r\kappa^2(\bSigma)$, we have
\begin{align*}
&\quad \mathbb{P}\left[ \left| \sum_{k'\neq k} \frac{\lambda_{k'}}{\lambda_k} V_{k'} \right| \leq \kappa(\bSigma) \right] \\
& \geq 1 - 2\exp\left( - \frac{\kappa^2(\bSigma)}{4r\kappa^2(\bSigma)+2\sqrt{2}\kappa(\bSigma)}\right) \\
& \geq 1 - \exp\left( -\frac{1}{8r }\right)  \geq  \frac{1}{9r}.
\end{align*} 
where the last inequality follows from $ \exp\left( -\frac{1}{8r }\right)\leq 1-\frac{1}{8r } + \frac{1}{128r^2} \leq 1-\frac{1}{9r}$. Combined with \eqref{proof_six}, we obtain $ \bu_k^H {\bJ}\bu_k \geq   (1-2p)e^{-\kappa(\bSigma)} /(9r)$.
\end{proof}

\section{Proof of Lemma~\ref{J_concentration}} \label{proof_J_concentration}

\begin{proof} We write
$$ \bJ_m - \bJ = \frac{1}{m}\sum_{i=1}^m \left[ z_i \bW_i -  \bJ \right]: = \frac{1}{m}\sum_{i=1}^m \bB_i,$$
where $\bB_i = z_i \bW_i -  \bJ$. To apply Lemma~\ref{matrix_bernstein}, we have $\mathbb{E}[\bB_i]=0$, and
\begin{align*}
\| \bB_i \| & \leq \| z_i (\ba_i\ba_i^H - \bb_i\bb_i^H )  \| + \|  {\bJ} \| \nonumber \\
& \leq  \|\ba_i\|_2^2 + \|\bb_i\|_2^2 +  1
\end{align*}
where the last inequality follows from $\| \bJ \|\leq (1-2p) $ due to Lemma~\ref{evd_expectedJ}. It is obvious that both $\|\ba_i\|_2^2$ and $\|\bb_i\|_2^2$ are chi-squared random variables with ``complex'' degrees of freedom $n$, then $\bB_i$ is sub-exponential with bounded sub-exponential norm. We have $\| \bB_i\|_{\psi_1}\leq Cn$ for some constant $C$. For the variance, we need to compute $\mathbb{E}[\bB_i^H \bB_i]$:
\begin{align*}
& \quad \mathbb{E}[\bB_i^H \bB_i] \\
& = \mathbb{E}\left[(z_i (\ba_i\ba_i^H - \bb_i\bb_i^H) -  \bJ)^H (z_i (\ba_i\ba_i^H - \bb_i\bb_i^H) - \bJ)\right] \\
& =  \mathbb{E}\left[(\ba_i\ba_i^H - \bb_i\bb_i^H) (\ba_i\ba_i^H - \bb_i\bb_i^H) \right] -  {\bJ}^H {\bJ} \\
& = 2 \mathbb{E}\left[\ba_i\ba_i^H \|\ba_i\|_2^2\right] -2 \bI -  {\bJ}^H {\bJ} = n\bI -  {\bJ}^H {\bJ},
\end{align*}
therefore $\sigma^2 =\left\| \sum_{i=1}^m\mathbb{E}[\bB_i^H\bB_i]\right\| \leq \sum_{i=1}^m \left\|\mathbb{E}[\bB_i^H\bB_i]\right\| = m\cdot \max\{ n , \|  {\bJ}^H{\bJ}\| \} = mn$. Applying Lemma~\ref{matrix_bernstein} we have,
\begin{align*}
\mathbb{P}\left[ \left\| \frac{1}{m} \sum_{i=1}^m \bB_i \right\| > \tau \right]  \leq 2n \exp\left( -\frac{\tau^2m}{2mn^2 + 2Cn \tau /3 }\right).
\end{align*}
Rearranging will conclude the proof.
\end{proof}

\section{Proof of Lemma~\ref{lemma:concentration}}\label{proof_lemma_concentration}
\begin{proof}
First, write 
$$\mathcal{T}\left(\bJ_m - {\bJ} \right)= \mathcal{T}\left(\frac{1}{m} \sum_{i=1}^{m}\left[z_i \bW_i \right] - {\bJ} \right)=  \frac{1}{m} \sum_{i=1}^{m} \bX_i,$$ 
where $\bX_i = \mathcal{T}\left(z_i \bW_i -\bJ \right)$ with $\mathbb{E}[\bX_i] = 0$. According to \cite[Lemma 5]{chen2013exact}, we have the event
$$\mathcal{E}_i =\left\{\|\mathcal{T}\left(\ba_i \ba_i^H\right)\| \leq c_1 \log^{\frac{3}{2}}(n)\right\}$$
holds with probability at least $1-n^{-10}$. Denote the event
$$ \mathcal{E} =\left\{ \forall 1\leq i\leq m: \|\mathcal{T}\left(\bW_i \right)\|\leq c_2 \log^{\frac{3}{2}}(n)\right\} ,$$
which holds with probability at least $1-(2m)\cdot n^{-10}$. Under the event $\cE$,
we can bound $\|\bX_i\|$ as
\begin{align*}
\|\bX_i\| &= \|\mathcal{T}\left(z_i\bW_i  -  \bJ \right)\|   \\
& \leq\|\mathcal{T}\left(z_i\bW_i \right)\|+\| \cT(\bJ) \|  \leq c_3\log^{\frac{3}{2}}(n),
\end{align*} 
where we used $\|\cT(\bJ)\|\leq \|\bJ\|\leq 1$. However, this conditioning will deteriorate the zero-mean property of $\bX_i$. Luckily, we will show the violation is small, and we can still bound the concentration in a desirable manner following a similar treatment in \cite[Appendix B]{candes2011probabilistic}. Denote the conditional expectation as $\bM = \mathbb{E}[\bX_i|\cE]$. We have
\begin{equation}\label{near_isometry}
\| \bM \|\leq \frac{c}{n^8}
\end{equation}
for some constant $c$, which will be shown at the end of the proof.
Also, conditioned on $\cE$, we have
\begin{align}
\left\|\sum_i\mathbb{E}\left[(\bX_i-\bM)^2| \cE \right] \right\|
&= m \left\| \mathbb{E}\left[(\bX_i-\bM)^2| \cE \right]\right\|\nonumber \\
&\leq c_1 m\cdot\left\|\mathbb{E}[\mathcal{T}\left(z_i\bW_i\right)^2 | \cE ]\right\| \nonumber \\
&\leq c_3 m\cdot \log^3(n) \nonumber
\end{align}
for some constant $c_3$. Applying the matrix Bernstein inequality in Lemma~\ref{matrix_bernstein}, we have that conditioned on $\cE$, as long as $m>c\log n$, with probability at least $1-n^{-9}$, we have
\begin{equation*}
\left\| \frac{1}{m}\sum_{i=1}^m\bX_i - \bM \right\| \leq \frac{\log^2 n}{\sqrt{m}}.
\end{equation*}
Finally, if $m$ further satisfies $m<cn$ for some $c$, we can have that with probability at least $1-n^{-9}$, 
\begin{align*} 
\left\| \frac{1}{m}\sum_{i=1}^m\bX_i \right\| & \leq \left\| \frac{1}{m}\sum_{i=1}^m\bX_i - \bM \right\| +\left\| \bM \right\| \leq c_2 \frac{\log^2 n}{\sqrt{m}}.
\end{align*}

The proof is complete after we prove \eqref{near_isometry}. Using the law of total expectation, we expand $\mathcal{T}(\bJ)$ as:
\begin{align}\label{total_expectation}
\mathcal{T}( {\bJ}) = \mathbb{E}\left[\mathcal{T}\left(z_i\bW_i\right)\right] \nonumber
&= \mathbb{E}\left[\mathcal{T}\left(z_i \bW_i\right)|\cE_i\right]\cdot \mathbb{P}\left(\cE_i\right)\\
&\quad + \mathbb{E}\left[\mathcal{T}\left(z_i \bW_i\right)I_{\cE_i^c}\right],
\end{align}
where $I_{\cE_i^c}$ is the indicator function of the event $\cE_i$. Our goal is to bound $\|\bM \|$ where $\bM = \mathbb{E}\left[\mathcal{T}\left(z_i\bW_i- {\bJ}\right)|\cE_i\right]$. After some basic transformation, \eqref{total_expectation} yields that    
\begin{align}
\|\bM\|&=\left\|\mathbb{E}\left[\mathcal{T}\left(z_i\bW_i\right)|\cE_i\right] -\mathcal{T}( {\bJ}) \right\|  \nonumber \\
&\leq \frac{1}{1-\mathbb{P}\left(\cE_i^c\right)}\left(\mathbb{P}\left(\cE_i^c\right) \| \mathcal{T}( {\bJ}) \|+\|\mathbb{E}\left[\mathcal{T}\left(z_i \bW_i\right)I_{\cE_i^c}\right]\|\right) \nonumber \\
& \leq  \frac{1}{1-\mathbb{P}\left(\cE_i^c\right)}\left(\mathbb{P}\left(\cE_i^c\right) +\|\mathbb{E}\left[\mathcal{T}\left(z_i \bW_i\right)I_{\cE_i^c}\right]\|\right).\label{boundM_intermediate}
\end{align}
Then it will be sufficient to bound $\left\|\mathbb{E}\left[\mathcal{T}\left(z_i \bW_i\right)I_{\cE_i^c}\right]\right\|$. By Jensen's inequality we have
\begin{align}
\|\mathbb{E}\left[\mathcal{T}\left(z_i \bW_i\right)I_{\cE_i^c}\right]\| 
&\leq \mathbb{E}\left[\|\mathcal{T}\left(z_i \bW_i\right)\|\cdot I_{\cE_i^c}\right] \nonumber\\
&\leq 2 \ \mathbb{E}\left[\|\mathcal{T}\left(\ba_i \ba_i^H\right)\|\cdot I_{\cE_i^c}\right]. \nonumber
\end{align}
Recall that according to \cite[Lemma 5]{chen2013exact},
$$\mathbb{P}\left(\left\|\mathcal{T}\left(\ba_i\ba_i^H \right)\right\|\geq c_1  \log^{\frac{3}{2}}(n)\right) \leq \frac{1}{n^{10}},$$ by letting $t = c_1 \log^{\frac{3}{2}}(n)$ and solving  for $n$ it can be alternatively described as
$$\mathbb{P}\left( \|\mathcal{T}\left(\ba_i \ba_i^H\right)\| \geq t   \right) \leq \exp\left(-c\cdot t^{\frac{2}{3}}\right):= f(t)$$ 
for some constant $t$. Following simple calculations, we have
\begin{align*}
&\quad \mathbb{E}\left[\|\mathcal{T}\left(\ba_i \ba_i^H\right)\|  I_{\mathcal{E}_i^c}\right] \\
& \leq c_1 \log^{\frac{3}{2}}(n)f\left(c_1  \log^{\frac{3}{2}}(n)\right)
+\int_{c_1  \log^{\frac{3}{2}}(n)}^{\infty} f(t)dt  \leq \frac{c}{n^8} 
\end{align*}
for some constant $c$ when $n$ is large enough. Thus we can bound $\|\bM\|$ as $\|\bM\|\leq c/n^8$ by plugging into \eqref{boundM_intermediate}.
\end{proof}

\bibliographystyle{IEEEtran}
\bibliography{bibfileCovOnebit,1bitToeplitz}

\end{document}